\documentclass[sigconf, nonacm]{acmart}

\usepackage{tikz}
\usepackage{amsmath}
\usepackage{subcaption}
\usepackage{listings}
\usepackage{graphics}

\usepackage{url}

\usepackage{breakurl}

\usepackage{multirow, makecell}

\newcommand{\remove}[1] {}

\makeatletter
\lst@Key{countblanklines}{true}[t]%
    {\lstKV@SetIf{#1}\lst@ifcountblanklines}
    
\lst@AddToHook{OnEmptyLine}{%
    \lst@ifnumberblanklines\else%
       \lst@ifcountblanklines\else%
         \advance\c@lstnumber-\@ne\relax%
       \fi%
    \fi}
\makeatother

\begin{document}

\title{Optimizing Inference Performance of Transformers on CPUs}

\author{Dave Dice}
\affiliation{%
  \institution{Oracle Labs}
   \city{Burlington, MA}
   \country{USA}
}
\email{dave.dice@oracle.com}

\author{Alex Kogan}
\affiliation{%
  \institution{Oracle Labs}
   \city{Burlington, MA}
   \country{USA}
}
\email{alex.kogan@oracle.com}

\thispagestyle{fancy}
\fancyfoot[C]{\vspace{0.5cm} \today \hspace{1mm} \textbullet \hspace{1mm} Copyright Oracle and or its affiliates}

\begin{abstract}
The Transformer architecture revolutionized the field of natural language processing (NLP).
Transformers-based models (e.g., BERT) power many important Web services, such as search, translation, question-answering, etc.
While enormous research attention is paid to the training of those models, relatively little efforts are made to improve their inference performance.
This paper comes to address this gap by presenting an empirical analysis of scalability and performance of inferencing a 
Transformer-based model on CPUs.
Focusing on the highly popular BERT model, we identify key components of the Transformer architecture where the bulk of the computation happens, 
and propose three optimizations to speed them up.
The optimizations are evaluated using the inference benchmark from HuggingFace, and are shown to achieve the speedup of up to x2.37.
The considered optimizations do not require any changes to the implementation of the models nor affect their accuracy.
\end{abstract}

\maketitle

\section{Introduction}

The introduction of the Transfomer architecture for deep neural networks (DNNs) by Vaswani et al.~\cite{VSP17} has literally transformed the field of NLP.
It happened just a few years ago (in 2017, to be exact), and since then the field has exploded with an enormous wave of Transfomer-based 
models achieving state-of-the-art, and often super-human, performance on many NLP tasks, which just recently have been
considered unrealistically difficult to solve. 
BERT~\cite{DCL19}, RoBERTa~\cite{LOG19}, ALBERT~\cite{LCG20}, Transformer-XL~\cite{DYY19} are only very few examples in the vast sea of published models~\cite{XWD20}.
As of today, Transfomer-based models, and BERT in particular, power many important Web services, such as search~\cite{Nay19, NYZ20}, translation
and text classification~\cite{LK20}.

The big premise of the Transfomer-based models is that they can be pre-trained on huge amounts of unlabeled data 
(such as all of Wikipedia or a book corpus), and later fine-tuned to a specific task (e.g., question-answering) using just a small amount of
labeled, domain-specific data.
To achieve high accuracy, those models feature millions (and, at times, billions) of parameters, and require long and expensive training.
As a result, numerous efforts have been made to optimize the training performance of those models~\cite{LCG20, YRH20, LOG19, GHL19}.
At the same time, and despite the vast deployment of those models in practice, far less attention is paid to inference performance.
Furthermore, among the efforts that do target inference performance of Transformer-based models, many consider
GPU or smartphone-based deployments~\cite{FYZ20, WBC19, WWC20, Zhu19}, even though in many practical settings the inference is done on 
small CPU-based systems~\cite{NYZ19, LK20}.

This paper comes to address this gap by presenting an empirical analysis of scalability and performance of inferencing 
Transfomer-based models on CPUs.
We identify the key component of the Transformer architecture where the bulk of the computation happens, namely, the matrix 
multiplication (matmul) operations, and propose three optimizations to speed them up.

The first optimization is based on the observation that the performance of the matmul operation is heavily impacted not only by
the shape (dimensions) of the source matrices and the available computing resources (the number of worker threads), but also
by whether (at least) one of those matrices is provided in a transposed form.
We propose a lightweight method to adaptively choose the appropriate form of source matrices for the inference, which results
in substantial performance improvement of the latter.
The second optimization stems from the observation that an invocation of matmul operations in deep learning (DL) frameworks 
incurs a significant sequential overhead, leading to the poor scalability of those operations.
We analyze the source of the overhead, and demonstrate how the scalability of matmul operations (and the overall inference performance)
can be improved by reducing (some portion of) that overhead.
Finally, the third optimization builds on the realization that while performant matmul operations are typically implemented by partitioning 
matrices into sub-matrices and carrying out the actual computation using highly optimized inner kernels~\cite{GG08}, the
partitioning itself might be suboptimal and not fully utilize parameters of the underlying hardware (such as cache capacity).
We show how choosing different parameters for matrix partitioning results in faster matmul operations.
We evaluate the efficacy of our optimizations using the industry-grade inference benchmark from HuggingFace~\cite{WDS20}. 

We note that prior work shows many factors impacting the inference performance of DNN models~\cite{WWW19}, including the choice of 
a DL framework, a math library, a thread pool library, availability of certain hardware features, such as the support
for SIMD (single instruction multiple data), etc.
To make our analysis feasible, we make several conscious choices when setting up our experimentation environment, focusing
on the inference performance of BERT implemented in the widely used Pytorch framework~\cite{pytorch} built with the state-of-the-art oneDNN math
library~\cite{onednn} (previously known as MKL-DNN) and run on an Intel Skylake processor-powered system (which supports AVX512 SIMD instructions).
While the chosen setup is significant, we validate the generality of many of our findings in other
variations of our setup such as with other Transformer-based models and math libraries.
We also note that despite the focus of our work being on NLP and Transformer-based models in particular, we believe our findings
extend to any model in which matmul operations consume a significant portion of the inference time.

The rest of the paper is organized as follows. We provide the relevant background on Transformers and BERT in Section~\ref{sec:background}.
The related work is discussed in Section~\ref{sec:related}. We describe our evaluation setup in Section~\ref{sec:setup} and provide
the analysis of the inference performance of BERT on CPUs in Section~\ref{sec:analysis}.
Based on this analysis, we describe three optimizations for inference performance in Section~\ref{sec:opts}.
Finally, we conclude in Section~\ref{sec:conclusions} with a discussion of the results and some of the future directions.

\section{Background: Transfomer and BERT}
\label{sec:background}
The Transformer architecture~\cite{VSP17} is composed of two stacks of identical layers; those stacks are called encoder and decoder.
For the purpose of this paper, we will focus on the encoder stack only, which is used exclusively in many actual Transformer-based models, including BERT.
In fact, we will note upfront that BERT's model architecture is almost identical to the Transformer encoder, only tweaking the number of layers,
the activation function, etc.~\cite{DCL19}. 
Also, we note that BERT itself has multiple configurations that differ in the various model hyper-parameters 
(e.g., the ``base'' configuration for BERT has $12$ layers while the ``large'' one has $24$).
Unless specified otherwise, when we say BERT in this paper, we refer to to its ``base'' configuration~\cite{DCL19}.

Each encoder layer has two sublayers, the first being a \emph{multi-head self-attention mechanism} and 
the second being a \emph{position-wise fully-connected feed-forward network}.
A \emph{residual connection} is employed around each of the sub-layers, followed by \emph{layer normalization}.

The attention mechanism is at the heart of the Transformer architecture. 
For the purpose of this paper we will focus on the actual computations performed by this mechanism;
the explanation of the intuition behind those computations can be found in many excellent sources~\cite{Ala18, harvard-nlp}, including in the original paper~\cite{VSP17}.
Specifically, the attention mechanism takes as an input three matrices Q, K and V and computes the output matrix:

\begin{equation}
Attn(Q, K, V) = softmax(\frac{QK^T}{\sqrt d_k})V
\label{eq:attn}
\end{equation}

\noindent where $d_k$ is the attention input dimension ($64$ for the BERT model).
As mentioned above, each self-attention sublayer includes multiple heads ($12$ for the BERT model). 
The computed function of this sublayer is given by the following expressions:

\begin{equation}
\begin{aligned}
MultiHead(Q,K,V) = Concat(head_1, ..., head_h)W^O\\
\textnormal{where}~head_i=Attn(QW_i^Q,KW_i^K,VW_i^V)
\end{aligned}
\end{equation}

\noindent where $W^O$, $W_i^Q$, $W_i^K$ and $W_i^V$ are parameter matrices. 
Overall, the computation of the multi-head self-attention requires 4 matrix multiplications to create input token projections
(the Q, K and V matrices) and the projection of the concatenated output of all the multiple heads. (We note that when 
Transformer is implemented in Pytorch, each of those multiplications are performed during the computation of the corresponding 
Linear modules.) In addition, two batched matrix multiplications are required to calculate the $Attn$ function in Equation~\ref{eq:attn}. 
Furthermore, the self-attention sublayer includes the invocation of softmax and layer normalization operations.

As for the fully-connected feed-forward sublayer, it consists of two linear transformations with an activation function in between:

\begin{equation}
\begin{aligned}
FFN(x)=Act(xW_1+b_1)W_2 + b_2
\end{aligned}
\end{equation}

\noindent where $W_1$, $b_1$, $W_2$ and $b_2$ are weight and bias matrices, respectively (which are model parameters, one set for each layer) and 
$Act$ is an activation function, such as \emph{gelu}~\cite{HG16}. 
While the inputs and outputs of the feed-forward sublayer have the same dimensions as the rest of the model 
($768$, in case of BERT), the inner-layer has a larger dimensionality ($3072$ for BERT).
It is easy to see that the computation of the feed-forward sublayer requires two matrix multiplication operations (carried by two Linear modules in Pytorch), 
as well as an activation function and a layer normalization operation.

\begin{figure*}[!ht]
\subfloat[][Seq. length 8]{\includegraphics[width=0.33\linewidth]{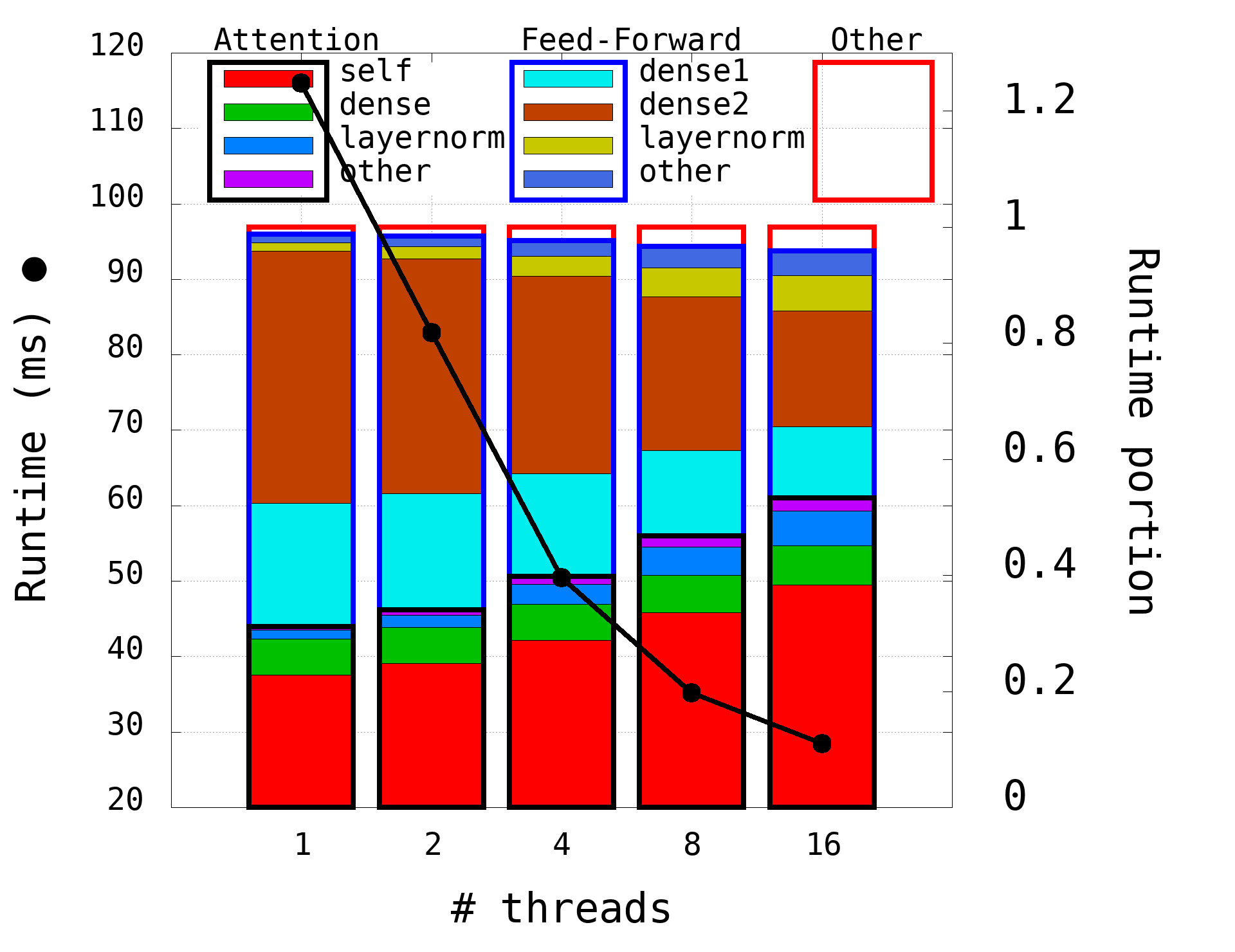}}
\subfloat[][Seq. length 64]{\includegraphics[width=0.33\linewidth]{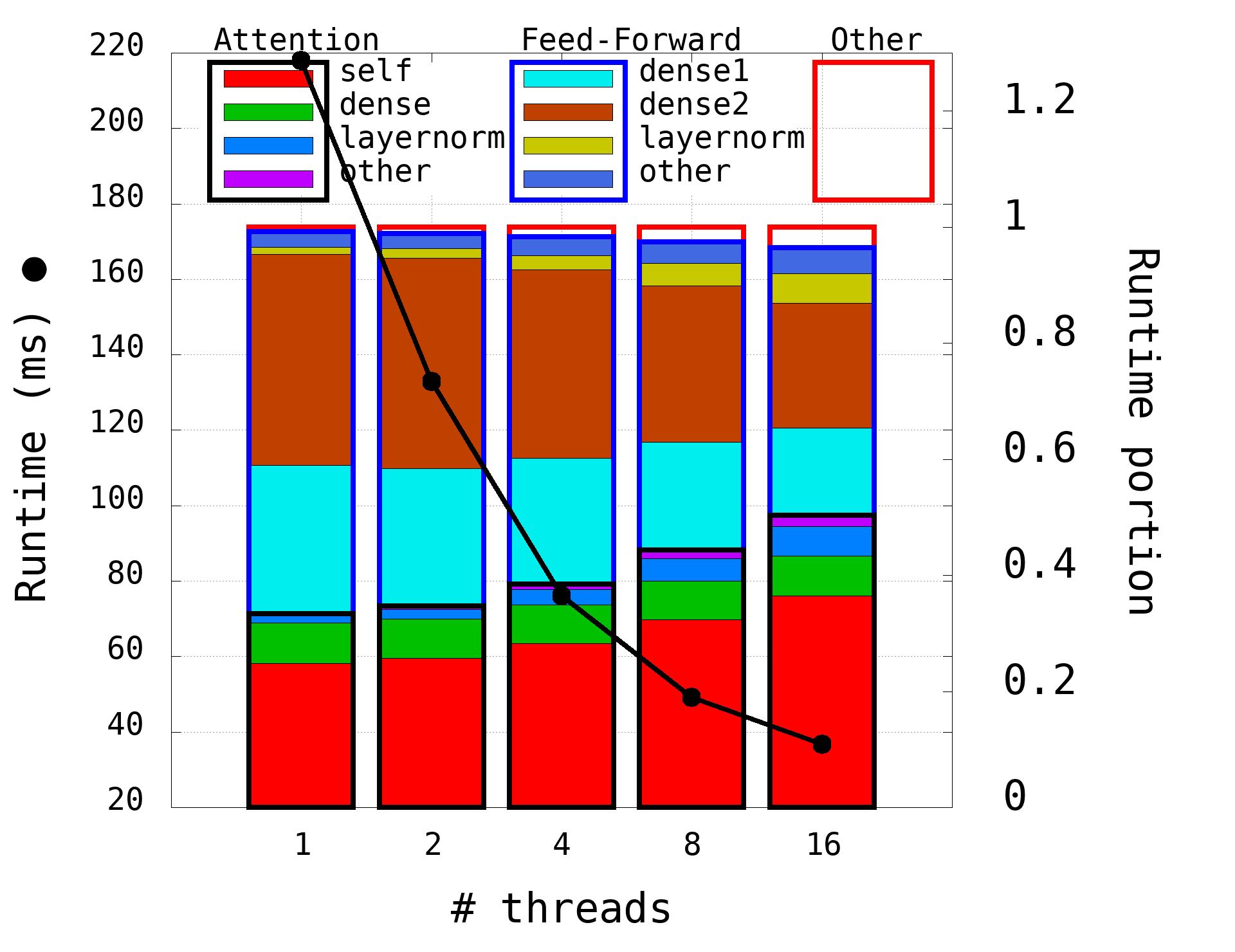}}
\subfloat[][Seq. length 384]{\includegraphics[width=0.33\linewidth]{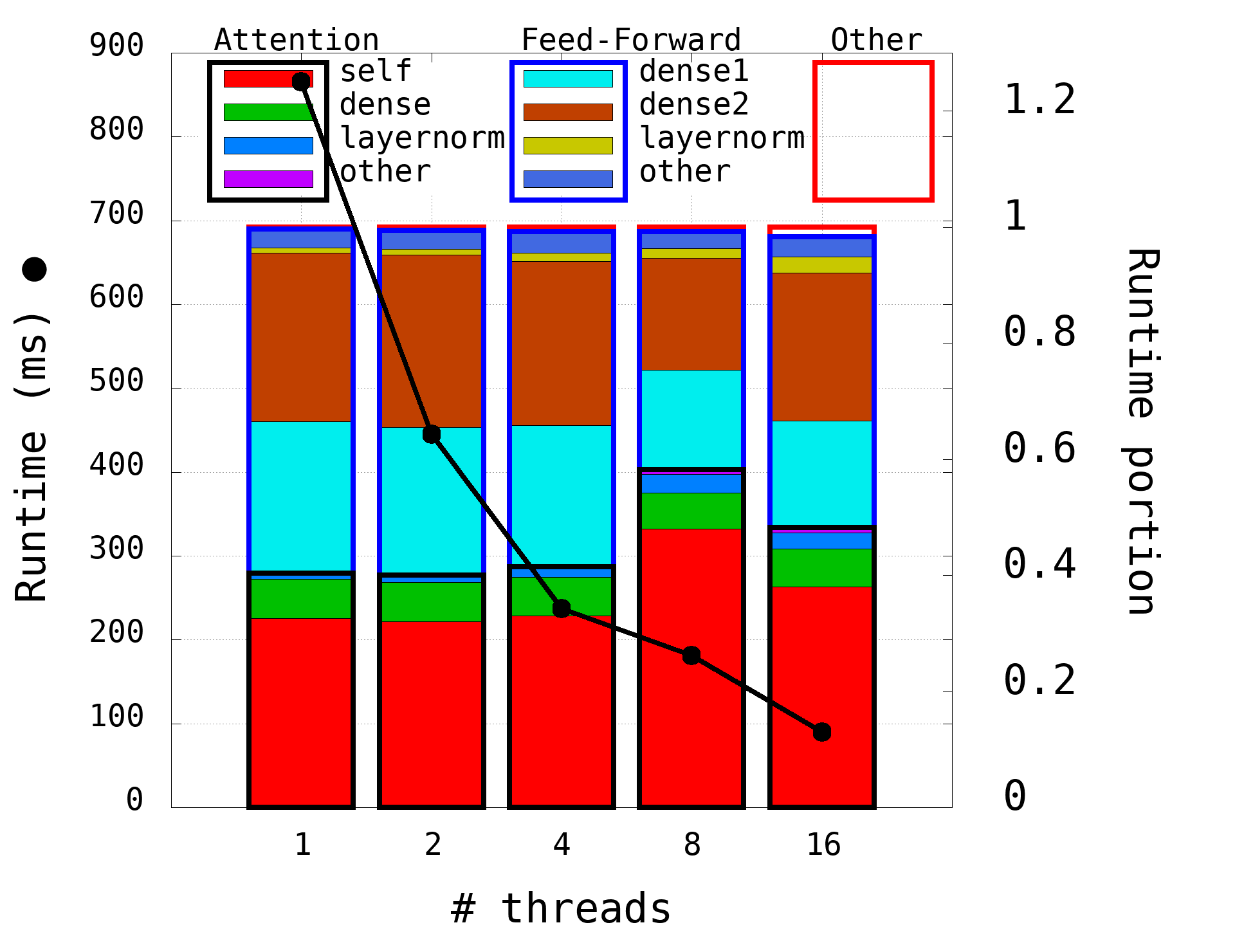}}
\caption{Performance breakdown for BERT by sub-layers and their components.}
\label{fig:bert-perf-breakdown-gross}
\end{figure*}

\section{Related Work}
\label{sec:related}
There is a relatively small body of work we are aware of on optimizing inference performance of NLP models on CPUs. 
Ning et al.~\cite{NYZ20} describe their effort on accelerating BERT with ONNX Runtime, an inference engine compatible with PyTorch and 
TensorFlow. The idea is to fuse multiple operations in the computation graph (e.g., matrix multiplication, layer normalization and gelu)
to reduce the amount of overhead (e.g., memory copying) in invoking each elementary computation individually. They also
experiment with reducing the number of layers in BERT sacrificing (some) accuracy for higher performance.
In general, we note that the operation fusion is a known technique for 
optimizing inference performance, and is orthogonal to the optimizations described in this paper.
Also, our techniques aim for performing the given inference computations faster, but without any change to the accuracy.

Wu et al.~\cite{NYZ19} describe another effort to optimize inference of BERT in Apache MXNet using the GluonNLP toolkit. They report 
on speedups achieved by using the MKL math library in MXNet as well as from quantizing the model for better performance with 
lower precision. We note that we use MKL as one of the baseline configurations in our analysis, while model quantization is, once 
again, orthogonal to the ideas discussed in this paper and may result in reduced accuracy.

There is an enormous effort on refining an/or replacing the attention mechanism with a more efficient alternative that requires
less computation and/or allows scaling for longer sentences, e.g.,~\cite{CTB20,ZGD20,WLM20, BPC20}.
While most of those efforts are primarily concerned with speeding up training, they help inference directly or indirectly as well.
Notably, one of the goals behind the \emph{knowledge distillation} effort~\cite{SDC19, SCG19, WWD20}, i.e., training a smaller model (student) to 
achieve a similar accuracy as a larger one (teacher), is reducing the inference latency.
Indeed, Le and Kaehler describe how they employ distillation with quantization to speedup their deployment of BERT on CPUs~\cite{LK20}.
We believe the optimizations described in this paper apply to most of such work.
In particular, we show the speedups achieved by our optimization for inferencing DistilBert~\cite{SDC19}, 
a popular model that uses knowledge distillation, are similar to those of BERT.

In a broader context, Liu at et.~\cite{LWY19} describe an approach called NeoCPU for optimizing CNN inference on CPUs. In addition to the common 
optimizations of operation fusion and inference simplification, NeoCPU manipulates the data layout flowing through the model
to minimize the overhead of transforming the data between various individual operations.
Fang et al.~\cite{FYZ20} present TurboTransformers, a GPU-based serving system for Transformer models.
They describe a number of optimizations targeting GPUs, such as memory management and batch reduction.
Optimizing inference of NLP models on GPUs has been also the motivation behind the work by Wang et. al~\cite{WWC20}.
At the same time, Wu et al.~\cite{WBC19} describe opportunities and design challenges in enabling machine learning inference
on smartphones and other edge platforms. 

\section{Evaluation Environment}
\label{sec:setup}
In this section, we describe the hardware and software setup for our experiments.
We ran the experiments on an Intel-based system featuring two Intel Xeon Platinum 8167M processors with 26 hyperthreaded cores each, and runs an Oracle Linux 7.8 OS.
To avoid any non-uniform memory access effects, we use the \texttt{numactl} utility to restrict all our experiments to executing on and allocating memory from 
one socket only.

On the software side, we use Pytorch v1.6, a popular DL framework.
We compile Pytorch in the default configuration, which means that it employs MKL as its default math library, but also includes support for oneDNN.
While MKL is a closed-source library, oneDNN is ``an open-source cross-platform performance library of basic building blocks for deep learning applications"~\cite{onednn},
and, unless stated otherwise, we use the latter in our experiments.

To invoke oneDNN bindings, one needs to convert a given model as well as the input tensors into the so-called ``mkldnn'' data format,
which dictates how data is laid out in memory~\cite{onednn-memory-format}.
We note that oneDNN bindings, however, are available for only a handful of DL operations, such as Linear and Convolution modules.
In practice, this means that for each such supported operation, Pytorch would seamlessly convert input tensors into the mkldnn format, apply the corresponding operation
and convert the output tensors back into the default (``dense'') format so they could be sent to other operations for which oneDNN bindings are not provided.
Such conversions do not come for free, however, and involve memory layout translations and memory copying.
To avoid this overhead, we extend the integration of oneDNN with Pytorch, adding the missing bindings for various operations 
invoked by a typical Transformer model, such as the layer normalization, softmax and gelu activation functions.
The extension comprises of a few hundred lines of C++ and Python code.
As we show in Section~\ref{sec:almo}, the resulting setup performs on-par with or better than the default Pytorch configuration (which employs MKL).

We use the popular Transformers Python package (v3.0.2) from HuggingFace, which provides a state-of-the-art implementation of numerous Transformer-based NLP models
implemented in Pytorch (and Tensorflow)~\cite{WDS20}. In addition, Transformers includes an easy-to-use inference benchmark, which we utilize heavily
for our experiments. Furthermore, we utilize mcbench~\cite{mcbench}, the open-sourced suite of microbenchmarks, which includes, among other things,
microbenchmarks for evaluating the performance of matrix multiplication operations when invoked directly through the C++ API of the corresponding math libraries.

\section{Inference Performance Analysis}
\label{sec:analysis}

\begin{figure*}[!ht]
\subfloat[][Seq. length 8]{\includegraphics[width=0.33\linewidth]{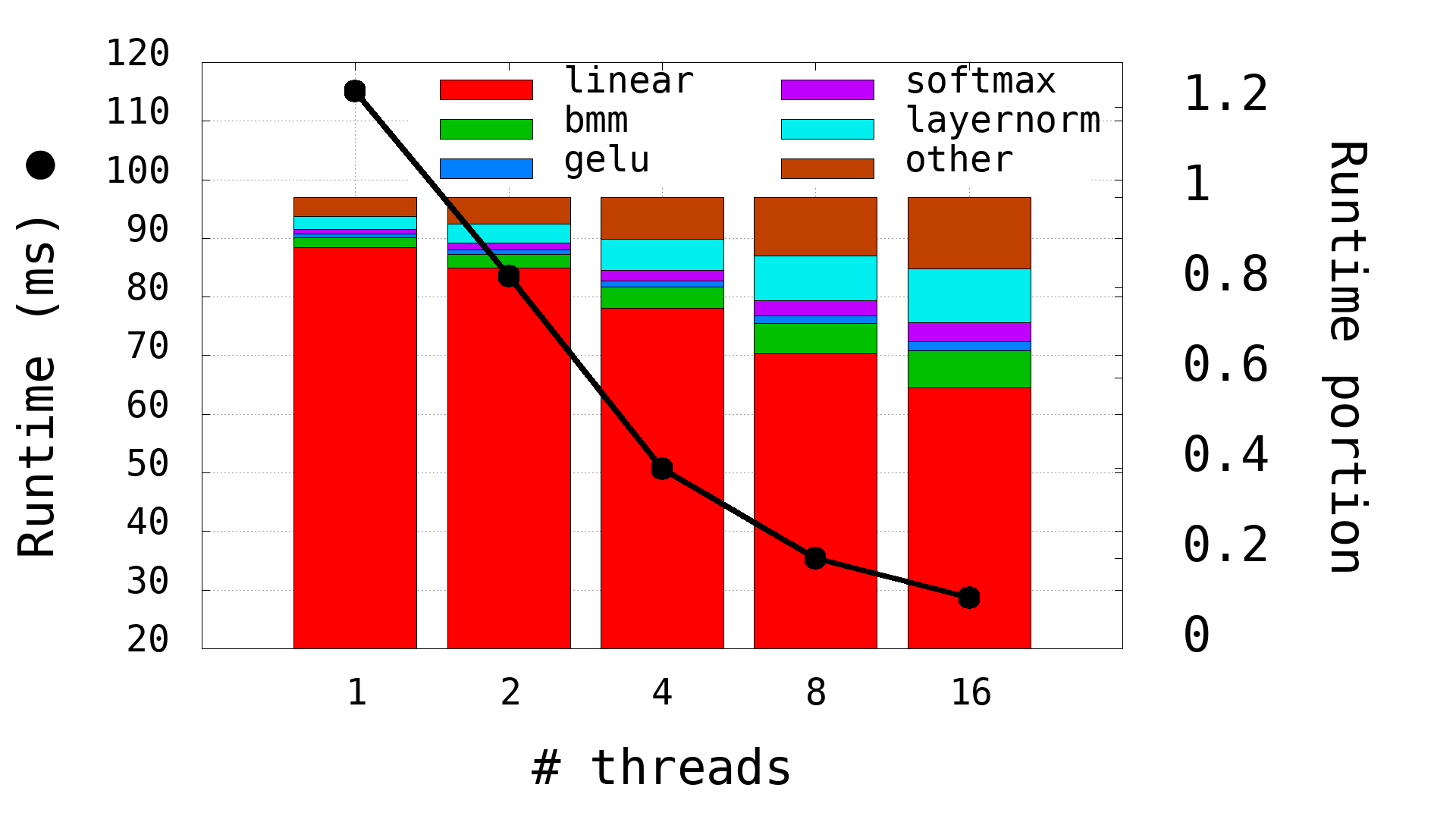}}
\subfloat[][Seq. length 64]{\includegraphics[width=0.33\linewidth]{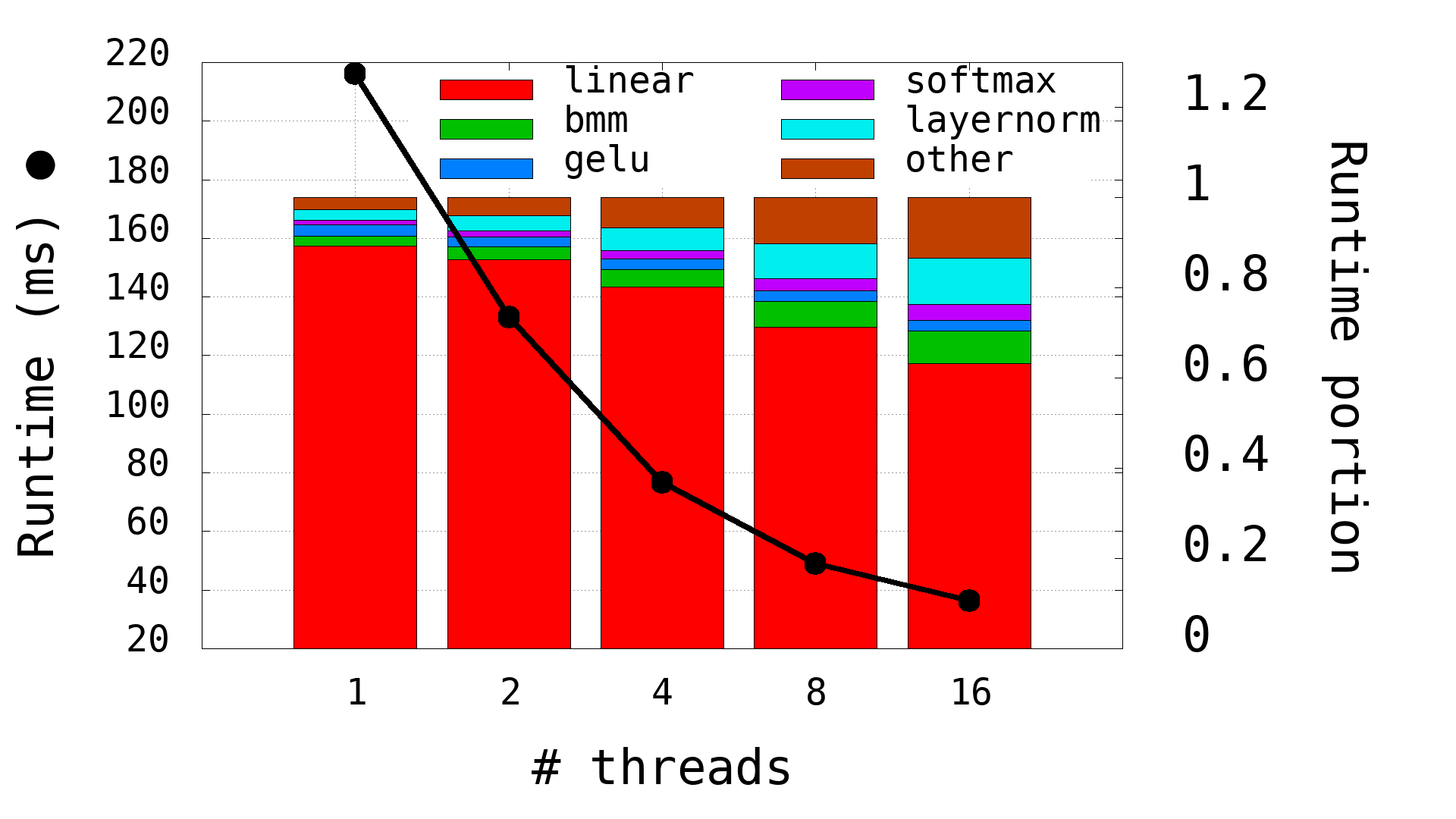}}
\subfloat[][Seq. length 384]{\includegraphics[width=0.33\linewidth]{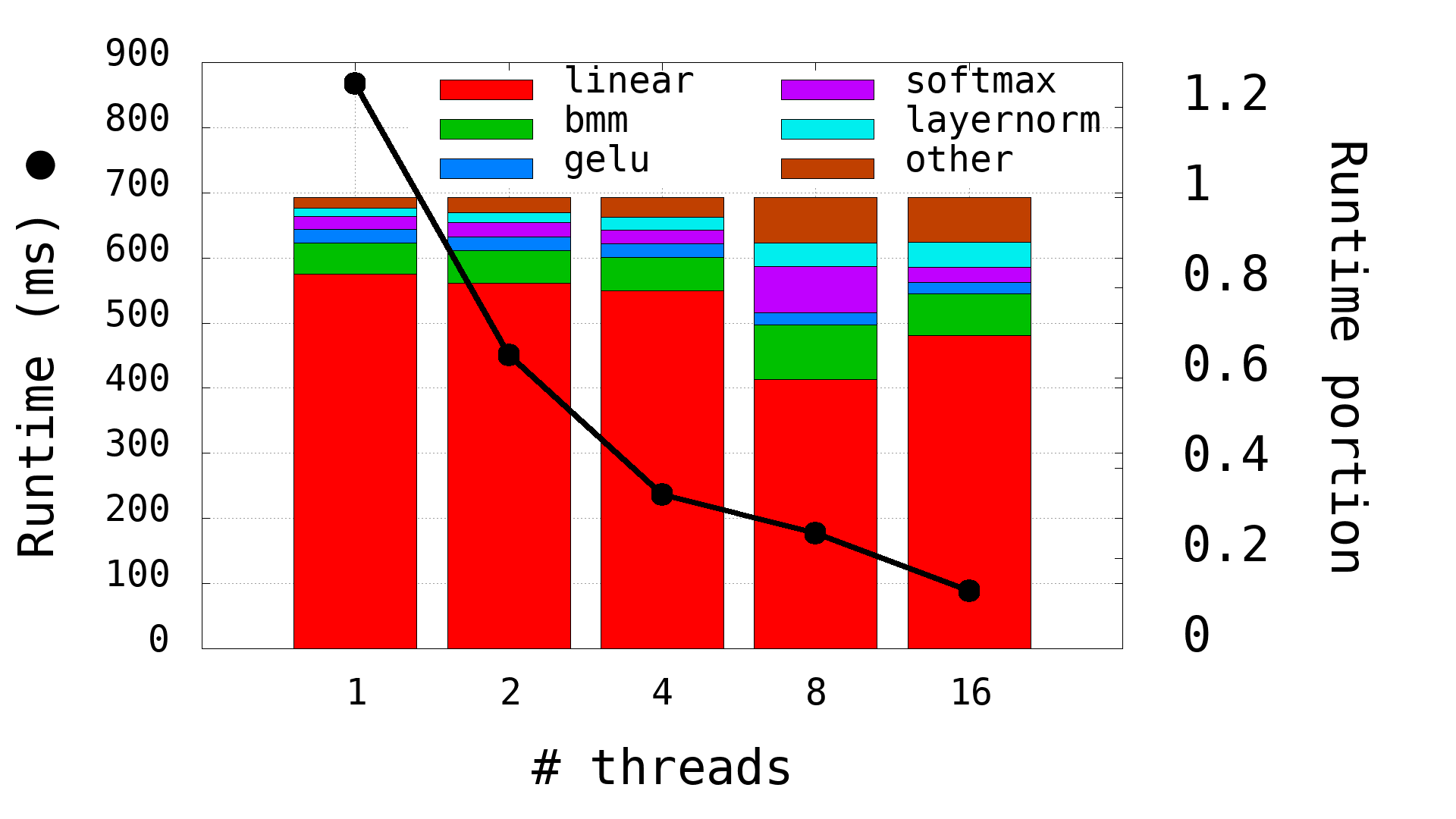}}
\caption{Performance breakdown for BERT by modules.}
\label{fig:bert-perf-breakdown}
\end{figure*}

We instrument the BERT model implemented in Transformers~\cite{WDS20} and collect timing information for various sub-layers (multi-head attention, feed-forward) and 
modules (Linear, Softmax, etc.) composing the model while executing the Transformers inference benchmark.
We experiment with various input sequence lengths and vary the number of threads from 1 to 16 
(by setting the \texttt{OMP\_NUM\_THREADS} environment variable).
We note that although our experimental machine has more than 16 cores, we deliberately decided to focus on smaller setups, 
as practical inference deployments typically include a small number of cores~\cite{NYZ19, LK20}.

Figure~\ref{fig:bert-perf-breakdown-gross} presents the inference latency as well as the breakdown of runtime spent in two main sub-layers, 
attention and feed-forward (along with the small portion of time not associated with any of the sub-layers, which consists mostly 
of input embedding and pooling of the resulting tensor; this time is denoted as a red box titled ``Other'').
In addition, we break the time in the two sub-layers into major components.
For the attention sub-layer, this is the time spent in self-attention (``self''), the linear projection (``dense''), the layer normalization (``layernorm'') and
the rest (``other'').
For the feed-forward sub-layer, this is the time spent in two linear projections (``dense1'' and ``dense2''), the layer normalization (``layernorm'') and
the rest (``other''), e.g., the gelu activation function.

Overall, we see that the feed-forward sub-layer typically consumes more time than the attention sub-layer.
Linear projections, which ultimately translate into matmul operations, are responsible for that.
We do not break down the attention sub-layer for better readability, yet the data shows that linear projections for the Q, K and V matrices 
consume the large share ($50$--$75\%$) of its time as well.
Along with that, the attention sub-layer does include other modules, e.g., softmax, tensor transpose and reshaping operations, etc.,
which explains why the time share of the attention sub-layer grows as we increase the number of threads.
Specifically, as we show in Section~\ref{sec:seq-overhead}, matmul operations, 
being carried out by carefully optimized math libraries (oneDNN, in this case), scale almost linearly with the number of threads.
At the same time, other operations (including layer norm, softmax and tensor reshaping) do not scale.
As such, their relative portion grows while the portion of time spent in matmul operations shrinks.
The portion of non-scalable operation is larger in the attention sub-layer, hence its weight grows with the number of threads.
This growth is more tamed when the input sequence is large (cf.~Figure~\ref{fig:bert-perf-breakdown-gross}~(c)) since the 
matmul operations are invoked with larger matrices, thus consuming a larger portion of time w.r.t.\ all other operations.

When examining the total runtime (the black curve in Figure~\ref{fig:bert-perf-breakdown-gross}),
we note that the overall scalability of the inference latency is relatively low, and depends on the input sequence length.
In particular, for sequences of 8 tokens, we achieve the speedup of only x3.3 when running with 16 threads versus 1 thread;
the speedup goes up to x9.7 for sequences of 384 tokens.
Better scalability with longer sequences is, once again, related to the scalability of matmul operations in math libraries and
the fact that larger sequences result in heavier matmul operations (with larger matrices) ---
reducing the time spent in matmul operations when the number
of threads increases has a larger effect on the overall scalability of the inference latency.

In Figure~\ref{fig:bert-perf-breakdown}, we present a different way to breakdown the inference runtime, by the time spent in various models.
While most module names are self-explanatory, we note that ``bmm'' stands for batched matrix multiplication, 
the operation at the heart of the multi-head attention mechanism 
(there are two of those operations per each attention layer in BERT); "other" stands for the time spent in computations 
not included in the specific modules, such as the time spent on transposing and reshaping tensors, input embedding, 
pooling of the resulting tensor, etc.

Not surprisingly, the vast majority of the inference time is spent in the Linear module, which in practice means matmul operations
(the Linear module applies a linear transformation to the incoming data, calculating a product of the input matrix with the stored weight matrix).
This concurs the results in Figure~\ref{fig:bert-perf-breakdown-gross}.
When summing up both ``linear'' and ``bmm'' runtime portions, the matmul operations consume 
between $66.2\%$ and $91.5\%$ of the total runtime. 
Note that this portion decreases as we increase the number of threads.
As explained above, this is because matmul operations are executed by a math library (oneDNN), 
and are highly optimized and scale almost linearly with the number of threads. 

\begin{figure*}[ht]
\subfloat[][\#threads=1 (oneDNN)]{\includegraphics[width=0.33\linewidth]{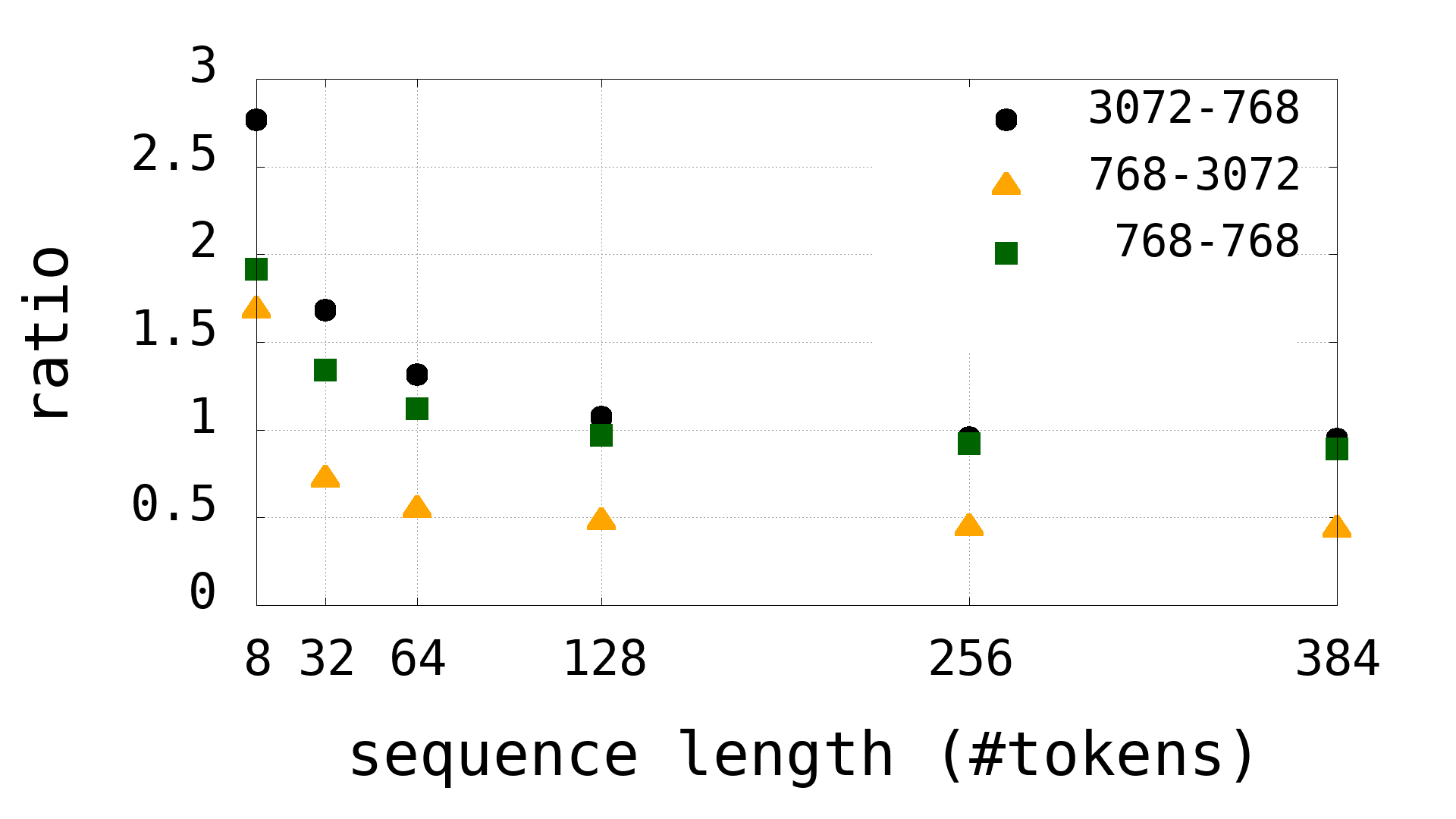}}
\subfloat[][\#threads=2 (oneDNN)]{\includegraphics[width=0.33\linewidth]{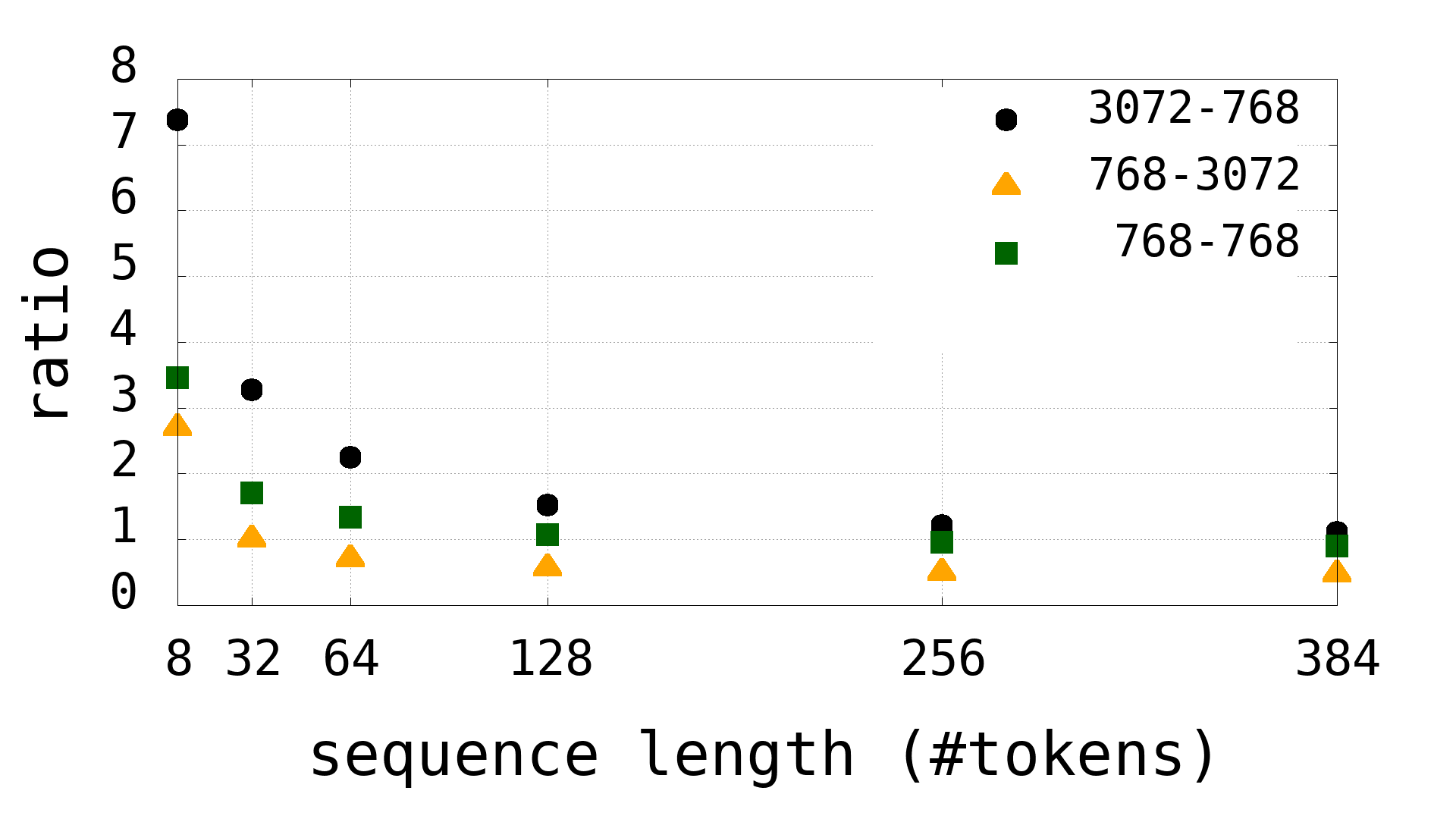}}
\subfloat[][\#threads=4 (oneDNN)]{\includegraphics[width=0.33\linewidth]{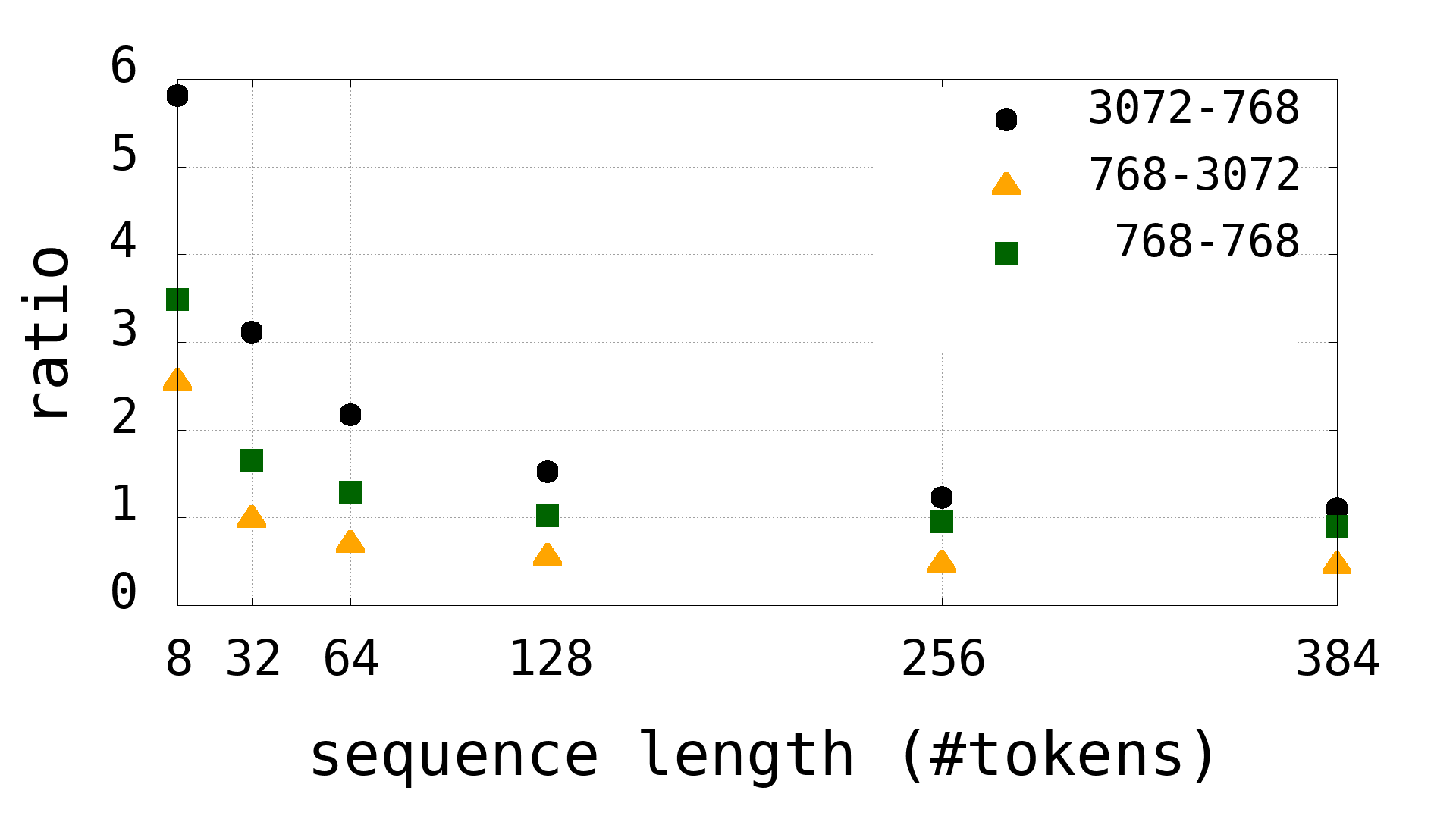}}\\
\subfloat[][\#threads=16 (oneDNN)]{\includegraphics[width=0.33\linewidth]{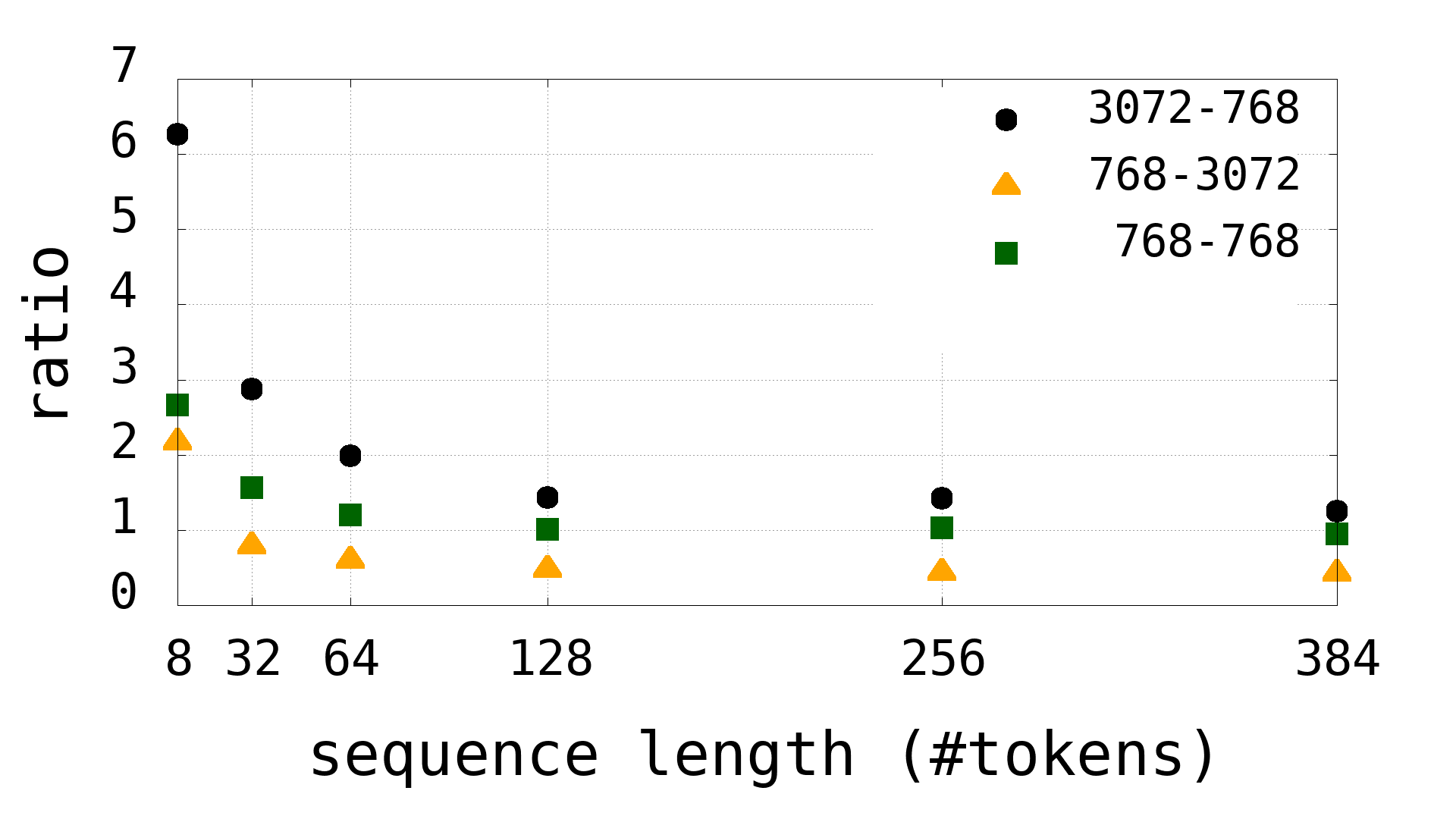}}
\subfloat[][\#threads=16 (MKL)]{\includegraphics[width=0.33\linewidth]{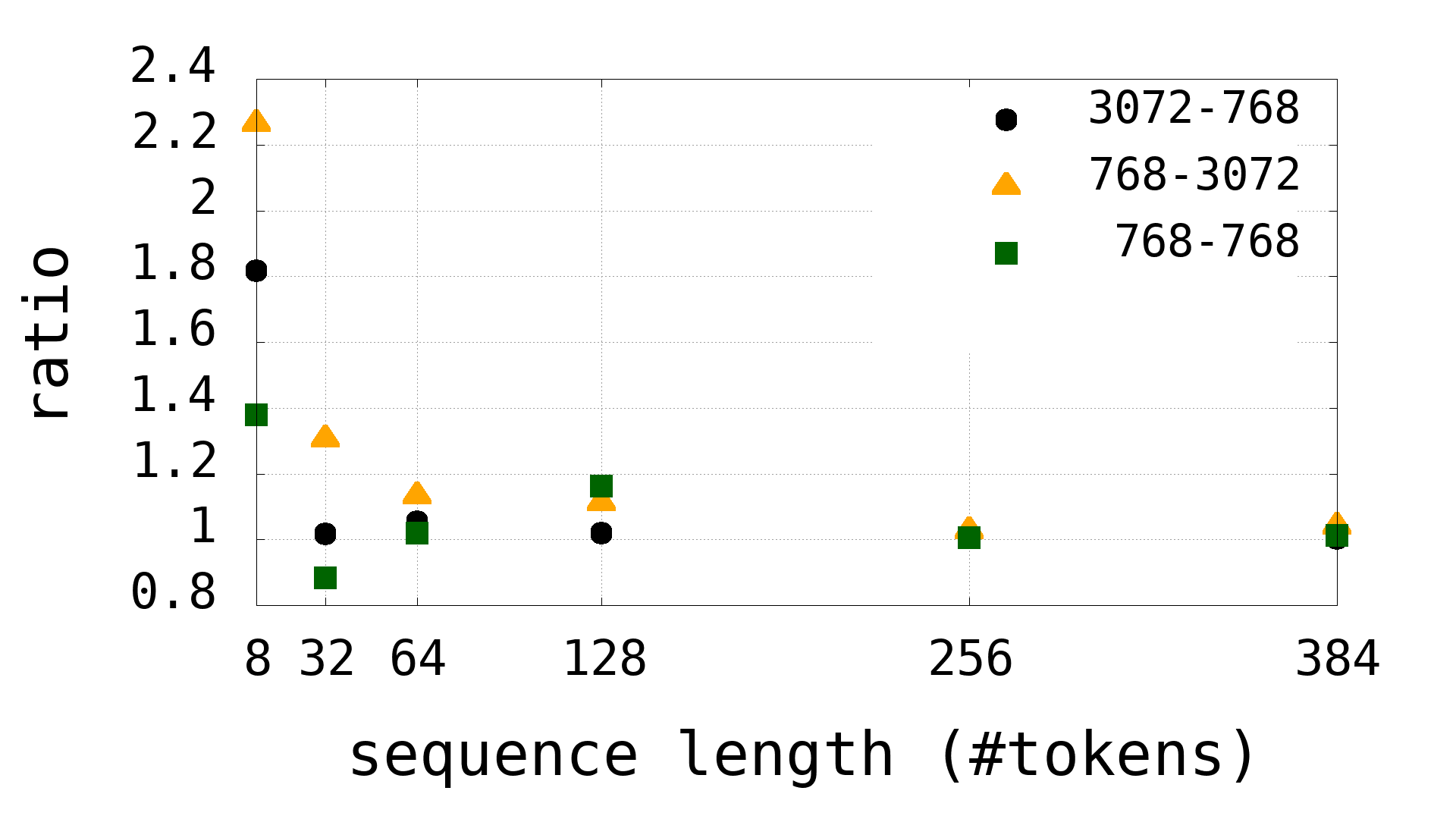}}
\subfloat[][\#threads=16 (OpenBLAS)]{\includegraphics[width=0.33\linewidth]{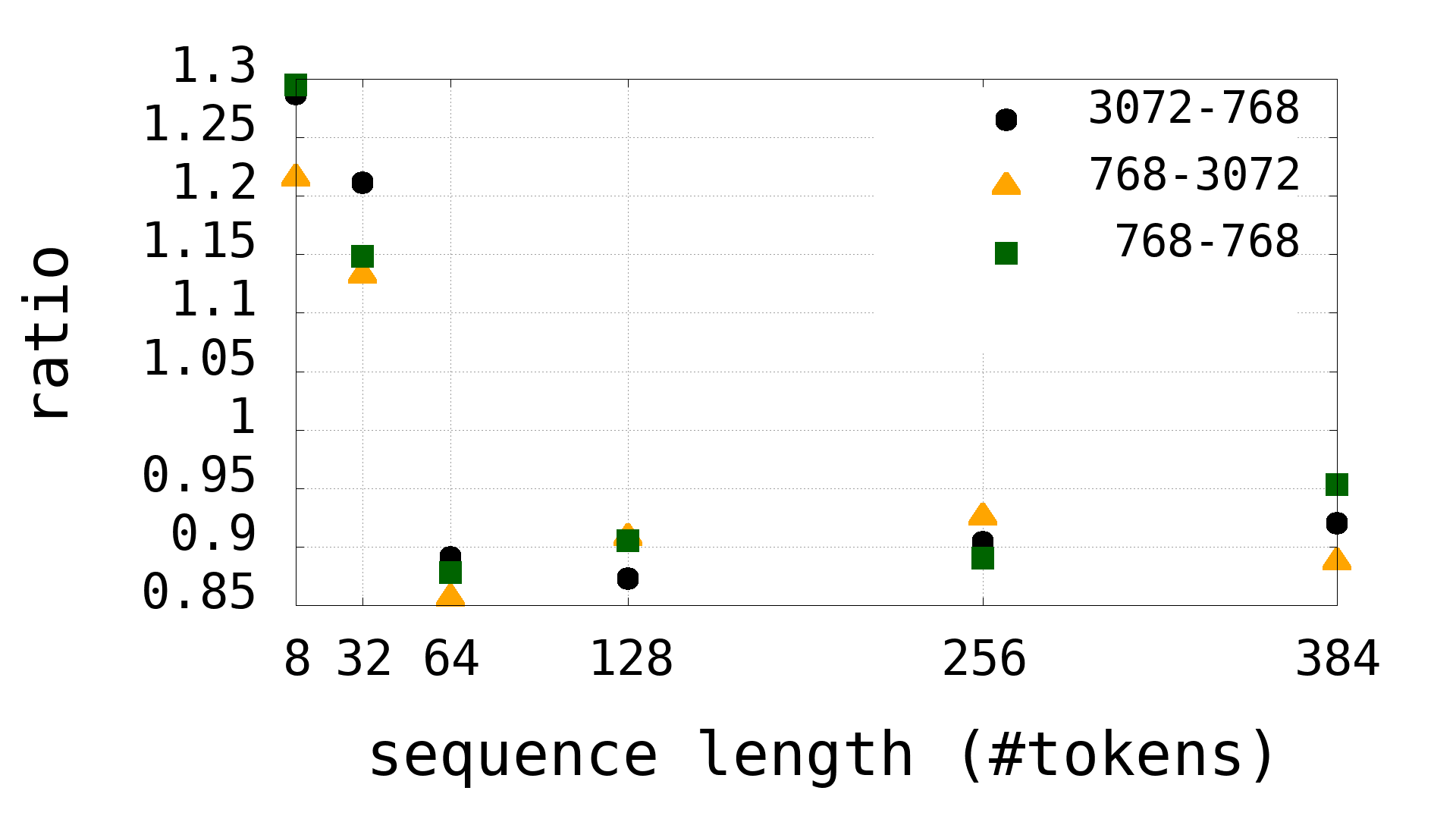}}
\caption{Matmul operation performance as a ratio between the time to multiply two non-transposed matrices and the time to multiply a non-transposed matrix by a transposed one.}
\label{fig:matmul-comp}
\end{figure*}

\section{Optimizing Inference Performance}
\label{sec:opts}
\subsection{Adaptive Linear Module}
\label{sec:almo}

The results in Section~\ref{sec:analysis} show that the key to improving inference performance on CPU lies in reducing 
the time spent in matmul operations. 

For the context, the API for a matmul operation allows invoking that operation on two source matrices A and B
(producing the destination matrix C=AB) s.t. each of the source matrices can be provided in the transposed form~\cite{onednn-blas}.
At the same time, the Pytorch Linear module stores the weight matrix in a transposed form, which means that,
during inference, the input matrix (A) is always non-transposed, while the weight matrix (B) is always transposed.
We believe the Pytorch designers made this choice (to transpose the weight matrix) to achieve better performance for 
the backward pass during training~\cite{pytorch-linear}, a concern which is not relevant for inference\footnote{We note
that in Tensorflow, the weight matrix is \emph{always} given in the normal form to the matmul operation.}.

Our experiments with the matmul microbenchmark from mcbench reveal an interesting observation. 
Figure~\ref{fig:matmul-comp} demonstrates the ratio 
between the time to compute matmul when both source matrices are non-transposed to the time to 
compute matmul when (only) the second source matrix is transposed. In other words, ratio~>~1 (ratio~<~1) corresponds to cases
in which the former (latter, respectively) method is faster. The shape of the second matrix (B) is represented by the name of the 
corresponding data series, while the shape of the first matrix (A) is given by the sequence length x first dimension of B.
Note that the chosen three shapes are not incidental, and they 
correspond to the shapes of weight matrices used in Linear modules of the BERT model.

More concretely, Figure~\ref{fig:matmul-comp}~(a)--(d) compare the performance of the matmul operation in oneDNN across different numbers of threads.
We see that for shorter sequences, multiplying the non-transposed matrices is almost always faster, and often results in substantial speedups.
For longer sequences, the picture is less clear -- one way of applying a mutmul operation is faster than the other for one shape but worse for another.
In general, the faster way of applying a matmul operation depends on the shape of the source matrices and the number of threads used.
We also confirmed that this observation is not unique to oneDNN, and is reproducible, at least to some extent, 
with other math libraries. Figure~\ref{fig:matmul-comp}~(e) and~(f) show the results obtained with MKL and OpenBLAS libraries, respectively 
(for the latter, we used the benchmark included in the library sources; for brevity, we include only the result for $16$ threads).

One may wonder about the reason for this performance difference. In oneDNN, much like in any math library for high-performance matmul calculation~\cite{GG08},
the matmul operation is coded in assembly, and each of the matmul variants (e.g., one with both source matrices in the normal
form vs.\ one in which the second matrix is transposed) results in a different code path, which generates different memory access patterns. 
Based on the profiling information produced by \texttt{perf},
we observe that given a certain configuration (i.e., the same source matrix shapes and the number of threads),
both variants have a similar number of L1 data cache misses, but the faster variant has a lower number of L3 cache accesses.
This suggests that one reason for performance difference might be the better utilization of L2 cache by one variant over the other.

\begin{table*}[ht]
\begin{center}
\resizebox{\linewidth}{!}{%
\begin{tabular}{c||ccc|ccc|ccccc}
\hline
\multirow{4}{*}{\#threads} & \multicolumn{10}{c}{sequence length}\\
\cline{2-12}
& \multicolumn{3}{c|}{8} & \multicolumn{3}{c|}{64} & \multicolumn{5}{c}{384}\\
\cline{2-12}
& \shortstack{onednn\\base} & \shortstack{onednn\\normal} & \shortstack{onednn\\almo} & \shortstack{onednn\\base} & \shortstack{onednn\\normal} & \shortstack{onednn\\almo} & \shortstack{onednn\\base} & \shortstack{onednn\\normal} & \shortstack{onednn\\almo} & \shortstack{mkl\\base} &  \shortstack{mkl\\script} \\
\cline{1-12}
1 & 115 & 82 & 79 $\pm$ 1 (x1.46) & 216 & 193 & 162 (x1.33) & 884 $\pm$ 31 & 1009 & 807 (x1.10) & 967 $\pm$ 15 & 861\\
2 & 83 & 50 & 50 $\pm$ 1 (x1.66) & 133 & 105 & 97 $\pm$ 1 (x1.37) & 471 $\pm$ 29 & 512 & 413 (x1.14) & 522 & 453\\
4 & 51 & 34 & 34 (x1.50) & 76 & 64 & 60 (x1.27) & 259 $\pm$ 27 & 285 & 220 (x1.18) & 302 $\pm$ 6 & 245\\
8 & 35 & 27 & 27 (x1.30) & 49 & 45 & 42 (x1.17) & 135 & 148 & 128 (x1.05) & 193 & 138 $\pm$ 2\\
16 & 28 & 24 & 24 (x1.17) & 36 & 34 & 32 (x1.12) & 96 $\pm$ 14 & 97 & 82 (x1.17) & 142 $\pm$ 2 & 83 $\pm$ 1\\

\hline
\end{tabular}%
}
\end{center}
\caption{BERT-base inference latency (ms), for various sequence lengths. The numbers in () show the speedup of \textbf{onednn-almo} over \textbf{onednn-base}.
The numbers after the $\pm$ sign specify the standard deviation when it is larger than 1\% of the mean.}
\label{table:bert-perf}
\end{table*}

Given the results in Figure~\ref{fig:matmul-comp}, we propose the following optimization for the Linear
module. 
Each Linear module is augmented with a \texttt{transposeFlags} array, specifying whether to use
a transposed version of the weights matrix for the forward pass (inference).
Entry $i$ of the array corresponds to the sequence length of $2^i$; the array has $10$ entries corresponding to the
maximal length of $512$ tokens.
When creating a Linear module with the given weights shape $[in,out]$, we generate random matrices with the shape $[2^i, in]$, 
for each $0 \leq i < 10 $,
and measure the time to perform a matmul operation when the weight matrix is transposed or not.
Based on the result, we set the corresponding entry \texttt{transposeFlags[i]}.
During the inference time, given the input of shape $[length, in]$, we calculate $s=\lfloor\log(length)\rfloor$,
and based on the flag in \texttt{transposeFlags[s]}, perform the matmul operation with either weight matrix transposed or not.

To avoid the overhead of transposing the weight matrix during inference, we keep both variants
of the weight matrix (transposed and non-transposed one).
This doubles the memory footprint of the Linear module.
While it might not be a concern on some CPU deployments, there are several ways to mitigate this drawback.
First, some shapes always prefer one form over the other, for all thread counts (e.g., 
the shape 3072-768 in Figure~\ref{fig:matmul-comp}~(a)--(d)). 
For this case, we can keep only the relevant variant of the weight matrix.
Second, the length of the input can be known prior to the deployment of an inference server, 
e.g., in a farm of inference servers, certain servers can be configured to handle input of a certain sequence length.
Once again, in this case we can keep only the relevant variant of the weight matrix.
Finally, if the input range is dynamic, one can store one variant of the weight matrix and transpose on-demand.
The selection of the stored variant can be also dynamic and tuned based on the actual input lengths seen during the runtime.
All those mitigation ideas are left for the future work.

We note that \texttt{transposeFlags} can be shared among Linear modules of the same shape.
We use a key-value map (dictionary) to store \texttt{transposeFlags} arrays where the $[in, out]$ tuple of corresponding Linear 
modules serves as a key.
Thus, when initializing the \texttt{transposeFlags} array, we query the dictionary first, and if such a shape has been already profiled, 
we reuse the resulting array, skipping the profiling phase for that Linear module.
For the BERT model, this optimization allows us to reduce the number of profiling phases from 73 
(6 Linear modules per each of the 12 self-attention layers plus one for the input embedding) to 3 (one per each different shape).
We emphasize that the profiling is run only once, during the initialization of the model (and its corresponding Linear modules), 
and is not invoked during inference.

Table~\ref{table:bert-perf} compares the performance of the HuggingFace inference benchmark run on top of several Pytorch variants
as described below.
Each experiment is run five times in the same configuration, and mean results are presented.
We also present the standard deviation for the cases where it was relatively high.
The variants we evaluate are \textbf{mkl-base}, which is the Pytorch version installed from \texttt{pip} (and uses MKL);
 \textbf{mkl-script}, which is the base version run in the torchscript mode (which creates ``a serializable and optimizable models from PyTorch code''~\cite{pytorch-jit}
 and therefore is a recommended mode for inference~\cite{pytorch-jit2}); \textbf{onednn-base}, which is the Pytorch version built from sources and uses oneDNN;
 \textbf{onednn-normal}, which is the onednn-base version in which the weight matrix is stored in a normal (non-transposed) shape;
 and \textbf{onednn-almo}, which is the onednn-base version with the adaptive Linear module optimization.
We note that the first two variants are included for reference only, to demonstrate that they perform mostly on-par with (and, at times, worse than) \textbf{onednn-base}.
Thus, we include them for one case only, for brevity.
We also note that the torchscript mode has a smaller impact when applied to oneDNN-based variants, shaving about $7$-$9$ ms from the reported 
latency in each case.
The qualitative comparison between oneDNN-based variants does not change, however 
(although, quantitatively, the torchscript mode leads to even larger speedups for the adaptive optimization).
Therefore, we do not include the torchscript mode results for those variants.

The improvements in the inference latency achieved by the adaptive Linear module optimization, as can be seen in Table~\ref{table:bert-perf}, 
correlate strongly with the results in Figure~\ref{fig:matmul-comp}. 
Specifically, higher speedups are achieved on shorter sequences and fewer number of threads, which are exactly the settings 
where the ratios depicted in Figure~\ref{fig:matmul-comp} are the highest.
We also note the need for adaptivity -- while \textbf{onednn-normal} performs well on shorter sequences, its performance suffers on longer ones,
which is exactly the settings in which, according to Figure~\ref{fig:matmul-comp}, multiplying the transposed weight matrix is faster.
The adaptive variant selects the correct shape in each case, performing on-par or better than the other two oneDNN-based variants.

We note that our findings extend to other Transformer-based models.
For instance, Tables~\ref{table:roberta-perf} and~\ref{table:distillbert-perf} present the results for RoBERTa and DistilBERT, respectively, 
in their ``base'' configurations, while Tables~\ref{table:bert-large-perf} presents the results for BERT-large, the larger version of the BERT model~\cite{DCL19}.
For brevity, we include only the inference latencies for \textbf{onednn-base} and \textbf{onednn-almo} variants.
While the results for RoBERTa and DistilBERT largely follow those for BERT-base, the results for BERT-large are slightly different. 
This is because BERT-large uses a different number of hidden units (1024 vs.\ 768 in other models we have considered), and thus operates with matrices of 
different dimensions.

\begin{table}[!t]
\begin{center}
\resizebox{\columnwidth}{!}{%
\begin{tabular}{c||cc|cc|cc}
\hline
\multirow{4}{*}{\#threads} & \multicolumn{6}{c}{sequence length}\\
\cline{2-7}
& \multicolumn{2}{c|}{8} & \multicolumn{2}{c|}{64} & \multicolumn{2}{c}{384}\\
\cline{2-7}
& \shortstack{onednn\\base} & \shortstack{onednn\\almo} & \shortstack{onednn\\base} & \shortstack{onednn\\almo} & \shortstack{onednn\\base} & \shortstack{onednn\\almo} \\
\cline{1-7}
\hline
1 & 116 & 79 (x1.47) & 216 & 162 (x1.33) & 860 & 809 (x1.06)\\
2 & 83 & 50 (x1.66) & 133 & 97 (x1.37) & 472 & 416 (x1.13)\\
4 & 50 & 35 (x1.43) & 76 & 60 (x1.27) & 250 & 221 (x1.13)\\
8 & 35 & 28 (x1.25) & 49 & 41 (x1.20) & 135 & 128 (x1.05)\\
16 & 29 & 24 (x1.21) & 36 & 32 (x1.12) & 112 & 82 (x1.37)\\

\end{tabular}%
}
\end{center}
\caption{RoBERTa inference latency (ms).}
\label{table:roberta-perf}
\end{table}

\begin{table}[!t]
\begin{center}
\resizebox{\columnwidth}{!}{%
\begin{tabular}{c||cc|cc|cc}
\hline
\multirow{4}{*}{\#threads} & \multicolumn{6}{c}{sequence length}\\
\cline{2-7}
& \multicolumn{2}{c|}{8} & \multicolumn{2}{c|}{64} & \multicolumn{2}{c}{384}\\
\cline{2-7}
& \shortstack{onednn\\base} & \shortstack{onednn\\almo} & \shortstack{onednn\\base} & \shortstack{onednn\\almo} & \shortstack{onednn\\base} & \shortstack{onednn\\almo} \\
\cline{1-7}
\hline
1 & 58 & 38 (x1.53) & 108 & 81 (x1.33) & 437 & 402 (x1.09)\\
2 & 41 & 24 (x1.71) & 67 & 49 (x1.37) & 226 & 210 (x1.08)\\
4 & 26 & 17 (x1.53) & 39 & 30 (x1.30) & 120 & 111 (x1.08)\\
8 & 18 & 14 (x1.29) & 25 & 21 (x1.19) & 69 & 65 (x1.06)\\
16 & 14 & 12 (x1.17) & 18 & 16 (x1.12) & 45 & 41 (x1.10)\\

\end{tabular}%
}
\end{center}
\caption{DistilBERT inference latency (ms).}
\label{table:distillbert-perf}
\end{table}

\begin{table}[!t]
\begin{center}
\resizebox{\columnwidth}{!}{%
\begin{tabular}{c||cc|cc|cc}
\hline
\multirow{4}{*}{\#threads} & \multicolumn{6}{c}{sequence length}\\
\cline{2-7}
& \multicolumn{2}{c|}{8} & \multicolumn{2}{c|}{64} & \multicolumn{2}{c}{384}\\
\cline{2-7}
& \shortstack{onednn\\base} & \shortstack{onednn\\almo} & \shortstack{onednn\\base} & \shortstack{onednn\\almo} & \shortstack{onednn\\base} & \shortstack{onednn\\almo} \\
\cline{1-7}
\hline
1 & 460 & 284 (x1.62) & 811 & 786 (x1.03) & 3134 & 3063 (x1.02)\\
2 & 379 & 163 (x2.33) & 558 & 475 (x1.17) & 1717 & 1726 (x0.99)\\
4 & 209 & 100 (x2.09) & 300 & 299 (x1.00) & 908 & 908 (x1.00)\\
8 & 124 & 68 (x1.82) & 171 & 203 (x0.84) & 506 & 501 (x1.01)\\
16 & 82 & 56 (x1.46) & 109 & 120 (x0.91) & 293 & 318 (x0.92)\\

\end{tabular}%
}
\end{center}
\caption{BERT-large inference latency (ms).}
\label{table:bert-large-perf}
\end{table}

For BERT-large and short sentences, \textbf{onednn-almo} achieves even more impressive gains over \textbf{onednn-base} compared to BERT-base, reaching the speedup of x2.33.
For longer sentences, however,  \textbf{onednn-almo} lags behind or performs on-par with \textbf{onednn-base}.
We identify the reason behind the performance regression as following: 
The baseline performance of matmul operations established during the profiling phase of 
the adaptive linear module optimization differs from the actual performance when the inference is executed.
In other words, during profiling, we establish that using the normal form of the weight matrix is faster than the transposed one, yet when we run inference, 
using weights in the normal form ends up being slower!
As we expand in Section~\ref{sec:matrix-partioning}, we hypothesize that this happens due to the poor fitting of matrix partitioning parameters in the math library 
(oneDNN) to hardware constraints, such as the L2 cache capacity.

\begin{figure*}[!ht]
\subfloat[][oneDNN matmul performance]{\includegraphics[width=0.5\linewidth]{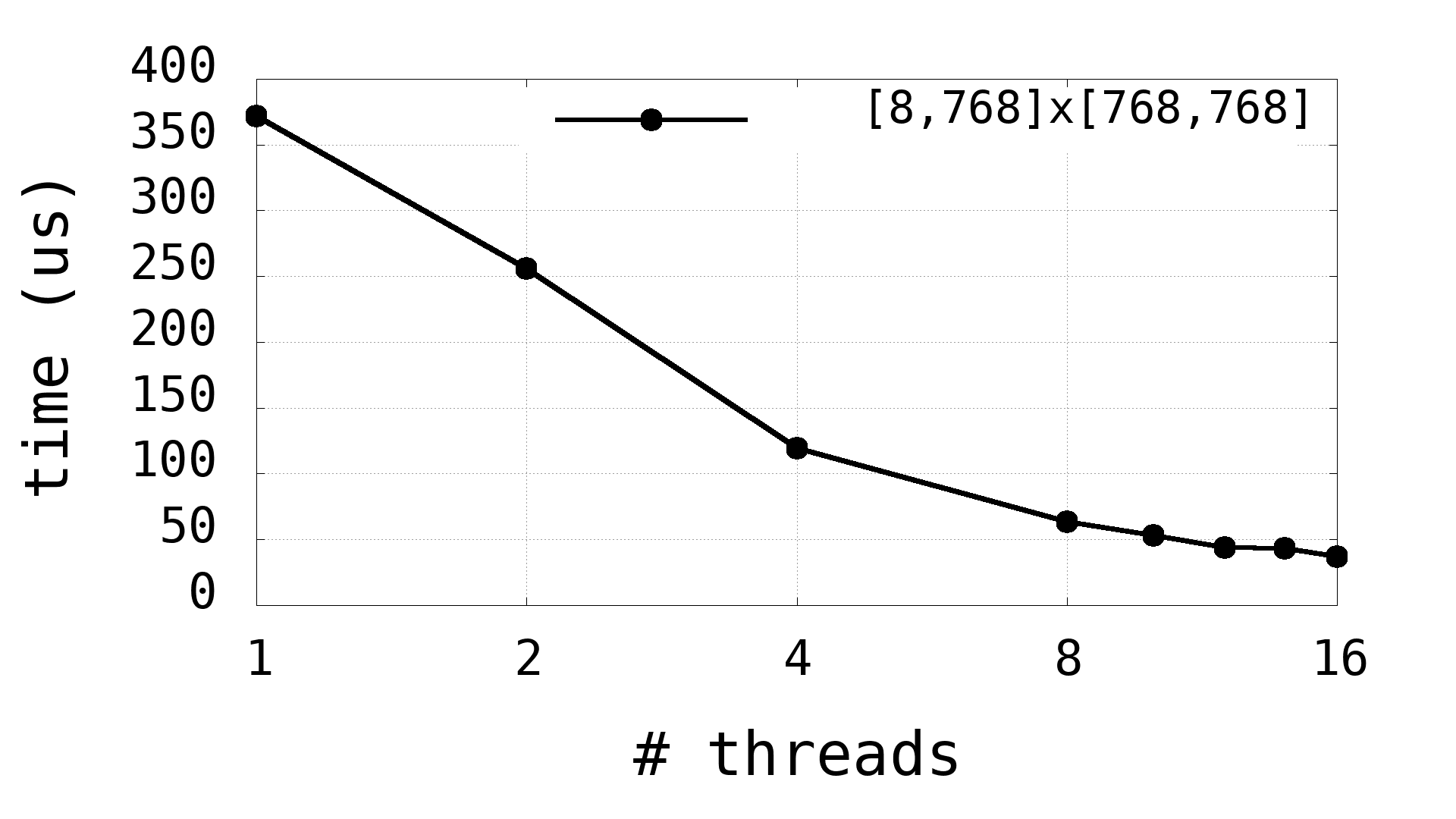}}
\subfloat[][Break down of Pytorch Linear module performance]{\includegraphics[width=0.5\linewidth]{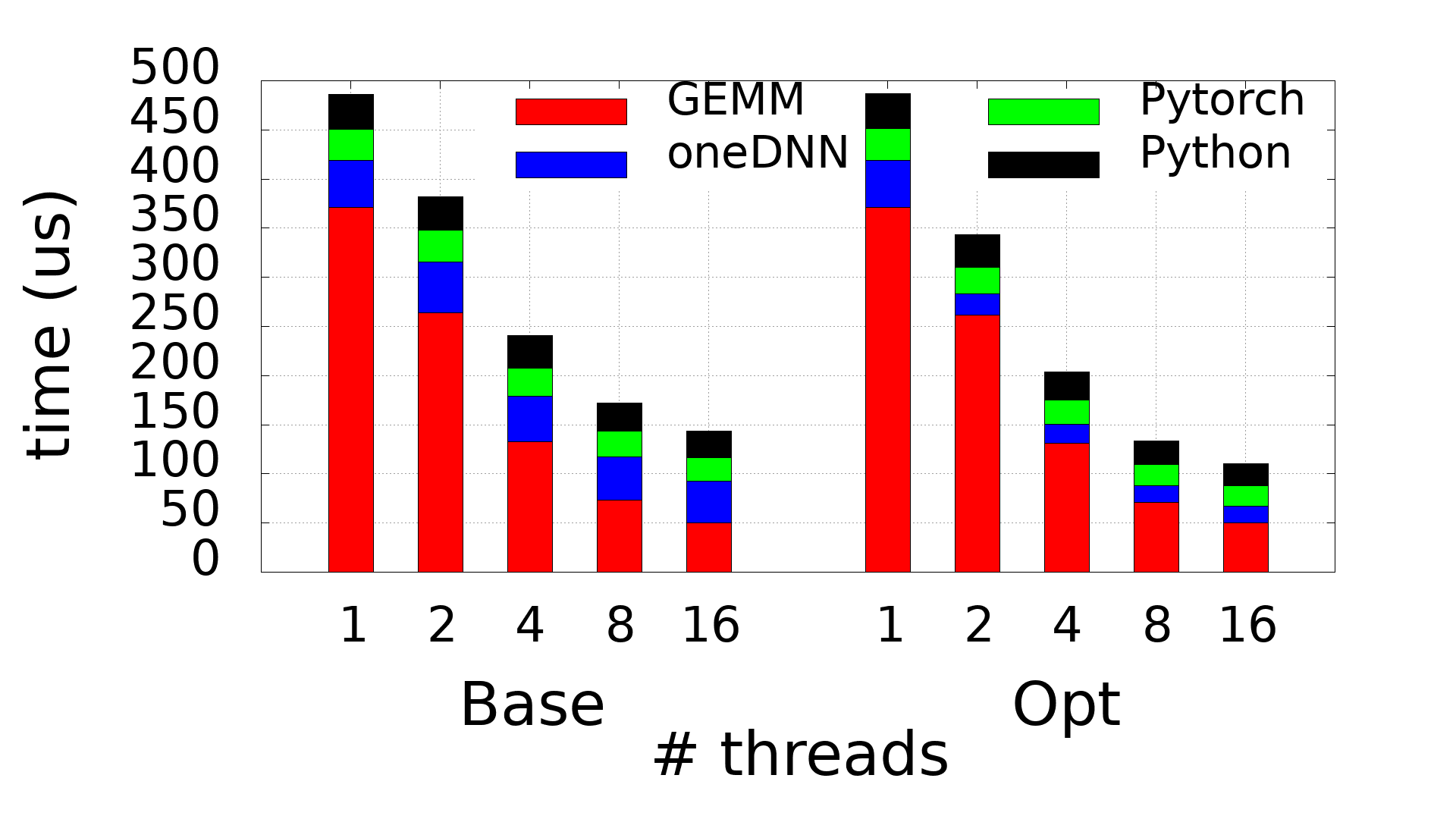}}
\caption{Matmul performance when invoked directly though oneDNN API and through the Pytorch Linear module.}
\label{fig:matmul-perf}
\end{figure*}

\subsection{Reducing Sequential Overhead}
\label{sec:seq-overhead}

The results in Table~\ref{table:bert-perf} (as well as in Figures~\ref{fig:bert-perf-breakdown-gross} and~\ref{fig:bert-perf-breakdown}) underline poor scalability of inference latency,
especially for shorter sequences. This is despite the fact that most of the inference time is spent in matmul operations 
(c.f.~Figure~\ref{fig:bert-perf-breakdown}), and those operations exhibit nearly linear scalability. The latter is demonstrated 
by the results from the mcbench microbenchmark shown in Figure~\ref{fig:matmul-perf}~(a), in which we measure the time to multiply
two matrices of the shape [8,768] and [768,768] as we vary the number of threads. 
(These shapes correspond to the matmul operation invoked by linear projections in the attention sublayer of the BERT model when the inference 
is performed on an input sequence of 8 tokens).

To shed more light on where the matmul operation cycles are spent during inference, we augmented Pytorch and oneDNN 
with timestamps. We ran the following simple code that employed the Linear module only (rather than a full-fledge model) 
and thus allowed to focus on the performance of matmul operations:

\begin{lstlisting}[language=Python, label=linear-microbenchmark,escapechar=|, commentstyle=\color{blue}]
import torch
from torch.utils import mkldnn as mkldnn_utils
net = torch.nn.Linear(768, 768)
net = mkldnn_utils.to_mkldnn(net)
seq = torch.rand(8, 768).to_mkldnn()
for i in range(0, 10000): net(seq)
\end{lstlisting}

Note that the forward path through the Linear module above invokes a matmul operation on two matrices of the same shape as
the ones used for the experiment in Figure~\ref{fig:matmul-perf}~(a).

With the collected profiling information, we break down the 
phases through which the invocation of the forward pass of the Linear module goes, 
separating the time spent in the Python interpreter, Pytorch dispatcher (that directs the call
to the oneDNN implementation of the Linear model), oneDNN dispatcher (that selects the appropriate low-level matmul function), and
finally, the matmul (aka general matrix multiply, or GEMM) function itself. 

The results are presented in the first (left) set of bars in Figure~\ref{fig:matmul-perf}~(b). 
They show that, indeed, the time in the GEMM function scales with the number of threads.
At the same time, the duration of the rest of the computation phases does not change as the number of threads increases, 
implying that matmul operations in Pytorch incur 
significant sequential overhead. This overhead becomes even more substantial when the matmul operation is applied on smaller matrices and/or
with a large number of threads. This, according to the Amdahl's law, explains the poor overall scalability of the matmul operation.

We focus on the oneDNN dispatcher as a target for reducing the sequential overhead. The dispatcher validates the input parameters,
identifies the capabilities of the underlying architecture (e.g., the type of supported AVX instructions, if any), the number of available threads, etc. 
Based on this information, it iterates over the list of available GEMM implementations and selects the first that is compatible with 
the given set of requirements for the matmul operation.
While this process is necessary for a correct behavior of oneDNN with arbitrary input matrices (or, in general, input parameters), we observe
that during inference, which constrains the set of possible inputs to a few specific shapes, only one particular GEMM function is called, at least when the 
number of threads is larger than one\footnote{The oneDNN dispatcher may choose a different function when it detects that only a single thread is available.}.
Thus, when more than one thread is used, we implement an optimization where the dispatching process is reduced to call that function directly, skipping 
the validation logic described above.

The result of this optimization is shown in the second (right) set of bars in Figure~\ref{fig:matmul-perf}~(b). 
We note that the time spent in the oneDNN dispatcher is reduced substantially, 
leading to increasing the speedup of x4.42 at 16 threads compared to a single thread (up from x3.4 without the optimization).
Overall, the speedup is still subpar to the one achieved with mcbench microbenchmark (cf.~Figure~\ref{fig:matmul-perf}~(a)) because the 
rest of the sequential overhead remains.

\begin{table}[!t]
\begin{center}
\resizebox{\columnwidth}{!}{%
\begin{tabular}{c||cc|cc|cc}
\hline
\multirow{4}{*}{\#threads} & \multicolumn{6}{c}{sequence length}\\
\cline{2-7}
& \multicolumn{2}{c|}{8} & \multicolumn{2}{c|}{64} & \multicolumn{2}{c}{384}\\
\cline{2-7}
&  \shortstack{onednn\\base} & \shortstack{onednn\\almo+sor} & \shortstack{onednn\\base} & \shortstack{onednn\\almo+sor} & \shortstack{onednn\\base} & \shortstack{onednn\\almo+sor} \\
\cline{1-7}
\hline
1 & 115 & 78 (x1.47) & 216 & 165 (x1.31) & 884 & 806 (x1.10)\\
2 & 83 & 48 (x1.73) & 133 & 94 (x1.41) & 471 & 412 (x1.14)\\
4 & 51 & 32 (x1.59) & 76 & 56 (x1.36) & 259 & 215 (x1.20)\\
8 & 35 & 25 (x1.40) & 49 & 39 (x1.26) & 135 & 123 (x1.10)\\
16 & 28 & 22 (x1.27) & 36 & 30 (x1.20) & 96 & 78 (x1.23)\\

\end{tabular}%
}
\end{center}
\caption{BERT-base inference latency (ms).}
\label{table:bert-perf2}
\end{table}

\begin{table*}[t]
\begin{center}
\resizebox{0.95\linewidth}{!}{%
\begin{tabular}{c||ccc|ccc|ccc}
\hline
\multirow{4}{*}{\#threads} & \multicolumn{6}{c}{sequence length}\\
\cline{2-10}
& \multicolumn{3}{c|}{8} & \multicolumn{3}{c|}{64} & \multicolumn{3}{c}{384}\\
\cline{2-10}
& \shortstack{onednn\\base} & \shortstack{onednn\\almo} & \shortstack{onednn\\almo+sor} & \shortstack{onednn\\base} & \shortstack{onednn\\almo} & \shortstack{onednn\\almo+sor} & \shortstack{onednn\\base} & \shortstack{onednn\\almo} & \shortstack{onednn\\almo+sor}  \\
\cline{1-10}
\hline
1 & 460 & 288 (x1.60) & 287 (x1.60) & 811 & 579 (x1.40) & 573 (x1.42) & 3134 & 3003 (x1.04) & 2954 $\pm$ 46 (x1.06)\\
2 & 379 & 165 (x2.30) & 160 (x2.37) & 558 & 317 (x1.76) & 311 (x1.79) & 1717 $\pm$ 22 & 1631 $\pm$ 39 (x1.05) & 1606 $\pm$ 40 (x1.07)\\
4 & 209 & 101 (x2.07) & 96 $\pm$ 1 (x2.18) & 300 & 191 (x1.57) & 186 $\pm$ 2 (x1.61) & 908 & 868 $\pm$ 15 (x1.05) & 881 (x1.03)\\
8 & 124 & 68 (x1.82) & 63 (x1.97) & 171 & 120 (x1.42) & 114 (x1.50) & 506 & 480 $\pm$ 7 (x1.05) & 467 $\pm$ 19 (x1.08)\\
16 & 82 $\pm$ 1 & 55 (x1.49) & 50 (x1.64) & 109 & 87 $\pm$ 1 (x1.25) & 82 (x1.33) & 293 $\pm$ 3 & 285 $\pm$ 12 (x1.03) & 275 $\pm$ 8 (x1.07)\\

\end{tabular}%
}
\end{center}
\caption{BERT-large inference latency (ms) with the modified matrix partitioning. The numbers in () show the speedup over \textbf{onednn-base}.
The numbers after the $\pm$ sign specify the standard deviation when it is larger than 1\% of the mean.}
\label{table:bert-large-patched}
\end{table*}

The effect of reducing the sequential overhead in matmul on the inference performance is shown in Table~\ref{table:bert-perf2}. 
Here we present the comparison between \textbf{onednn-base} and \textbf{onednn-almo+sor}, where the latter is the 
onednn-almo version with the sequential overhead reduction optimization described in this section applied. 
The new optimization shaves another 3--12\% from the inference time, with larger gains recorded at larger thread counts and/or smaller sequence lengths.
This is expected since those are the settings where the sequential overhead has the most relative impact on the duration of matmul operations.

\subsection{Modifying Matrix Partitioning}
\label{sec:matrix-partioning}

High-performance math libraries, including oneDNN, perform matrix multiplication by partitioning the arguments into sub-matrices of certain shape, which
are then given to assembly-coded inner-kernels~\cite{GG08}.
This design aims to amortize the cost of moving data across adjacent memory layers, all while taking advantage of carefully engineered inner-kernels.
Hence, on a high level, the matrix multiplication operation can be expressed as the following triple-nested loop:

\begin{lstlisting}[language=C, label=matmul-op, escapeinside={(*}{*)}, commentstyle=\color{blue}]
for (p = 0; p < sizeK; p+=BK)
   for (i = 0; i < sizeM; i+=BM)
      for (j= 0; j < sizeN; j+=BN) 
         (*$C_{ij}$*)+=(*$A_{ip}$*)(*$B_{pj}$*)
\end{lstlisting}

Various considerations take place when deciding on how to partition the matrices (i.e., set BK, BM and BN above), 
including the size of the caches and TLB (translation-look-aside buffers), the shape and layout of source matrices, etc.~\cite{GG08}.
Yet, while some of those parameters are clearly hardware dependent, the oneDNN implementation uses a set of constants to control the partitioning\footnote{e.g., 
see \texttt{sgemm\_nocopy\_driver()} in \url{https://github.com/oneapi-src/oneDNN/blob/63c8b5ce84b0be266d1edad0420390f2e131cb29/src/cpu/x64/gemm/f32/jit\_avx512\_common\_gemm\_f32.cpp\#L1808-L1816}}.
We strongly believe that the regressions reported for BERT-large in Section~\ref{sec:almo} are the results of excessively conservative fitting of those parameters to the actual hardware.

As evidence, we reduce one of the parameters (BK, from $384$ to $64$\footnote{We note that while we tried a few other settings for matrix partitioning,
a comprehensive sensitivity analysis of partitioning parameters is a part of the future work.}) so that, effectively, the matrix multiplication is carried by a larger number of iterations of the outermost loop,
where each inner-kernel is activated on smaller sub-matrices that are more likely to fit into cache and, in general, reduce the amount of cache misses.
This has a highly positive effect on the inference performance, as demonstrated in Table~\ref{table:bert-large-patched} with the results of the BERT-Large model, which shows significant 
gains for \textbf{onednn-almo} over \textbf{onednn-base} across most sequence lengths and thread counts.
Performance counters (reported by \texttt{perf}) show that with the patched version of oneDNN, the number of last-level cache (LLC) misses is reduced.
Also, while both (patched and non-patched) versions report a similar number of instructions, the patched version uses significantly less cycles, yielding a higher
IPC (instruction per cycle) ratio.

\begin{table}[t]
\begin{center}
\resizebox{\linewidth}{!}{%
\begin{tabular}{c||cc|cc|cc}
\hline
\multirow{4}{*}{\#threads} & \multicolumn{6}{c}{sequence length}\\
\cline{2-7}
& \multicolumn{2}{c|}{8} & \multicolumn{2}{c|}{64} & \multicolumn{2}{c}{384}\\
\cline{2-7}
& \shortstack{onednn\\base} & \shortstack{onednn\\almo+sor} & \shortstack{onednn\\base} & \shortstack{onednn\\almo+sor} & \shortstack{onednn\\base} & \shortstack{onednn\\almo+sor}  \\
\cline{1-7}
\hline
1 & 115 & 80 (x1.44) & 216 & 150 (x1.44) & 884 & 815 (x1.08)\\
2 & 83 & 48 (x1.73) & 133 & 84 (x1.58) & 471 & 421 (x1.12)\\
4 & 51 & 32 (x1.59) & 76 & 53 (x1.43) & 259 & 223 (x1.16)\\
8 & 35 & 24 (x1.46) & 49 & 37 (x1.32) & 135 & 128 (x1.05)\\
16 & 28 & 22 (x1.27) & 36 & 30 (x1.20) & 96 & 81 (x1.19)\\

\end{tabular}%
}
\end{center}
\caption{BERT-base inference latency (ms) with the modified matrix partitioning.}
\label{table:bert-base-patched}
\end{table}

The results for BERT-base with the modified matrix partitioning are given in Table~\ref{table:bert-base-patched}.
They show that reducing the size of matrix partitions has a favorable effect on this model as well.

\section{Discussion}
\label{sec:conclusions}

In this paper we present the analysis of the inference performance for BERT, one of the most prominent models for NLP based on the Transfomer architecture, on CPU-based systems.
The analysis demonstrates clearly that the way to speeding up inference lies through the optimization of the matmul operation.
Based on this observation, we investigate three optimizations for speeding up matmul operations, which collectively lead to the inference speedup of up to x2.37 
for Transfomer-based models over established baselines.
The optimizations do not require any changes to the implementation of those models, and they do not affect their accuracy.
We further note that while the focus of our work has been the Transfomer architecture, our results are applicable to any machine learning model in which matmul operations
consume a significant portion of the inference time.

Our work underscores the importance of the operation fusion as a technique for optimizing computation during inference~\cite{NYZ20, LWY19}.
Such fusion would reduce the amount of sequential overhead in invoking individual operations (cf.~Figure~\ref{fig:matmul-perf}~(b)) and, in general, 
bring the scalability of high-level operations, such as the Linear module computation, closer to their low-level counterpart (cf.~Figure~\ref{fig:matmul-perf}~(a)).
\remove{
We note that one of our optimizations for reducing the sequential overhead takes step in that direction.
However, it still leaves lots of performance on the table by not fusing the low level operations together and eliminating parts of the computation.
Optimizing further the self-attention operation through the expansion of the math library with more efficient, fused operations  is left for the future work.
}
Furthermore, our work demonstrates that tuning matrix partitioning can lead to substantial matmul speedups.
An adaptive approach similar to the one discussed in Section~\ref{sec:almo}, but applied to the matrix partitioning parameters, might be warranted.
 
Another related future direction is scaling primitive operations beyond matmul. While matmul is responsible for the lion's share of the 
inference time, the portion of other operations grows as the number of threads increases. For instance, for short sequences, 
the share of the time spent in the layer normalization operation grows from $2.9\%$ for 1 thread to $12\%$ for 16 threads (cf.~Figure~\ref{fig:bert-perf-breakdown}). 
Parallelizing those operations and fusing them with matmul should provide further improvement to the inference performance.
 
\bibliographystyle{plain}
\bibliography{refs}

\begin{thebibliography}{10}

\bibitem{Ala18}
Jay Alammar.
\newblock {The Illustrated Transformer}.
\newblock \url{http://jalammar.github.io/illustrated-transformer/}.
\newblock Accessed: 01-07-21.

\bibitem{BPC20}
Iz~Beltagy, Matthew~E. Peters, and Arman Cohan.
\newblock Longformer: The long-document transformer.
\newblock {\em CoRR}, abs/2004.05150, 2020.

\bibitem{CTB20}
Qingqing Cao, Harsh Trivedi, Aruna Balasubramanian, and Niranjan
  Balasubramanian.
\newblock Deformer: Decomposing pre-trained transformers for faster question
  answering.
\newblock In {\em Proc. of Conference of the Association for Computational
  Linguistics (ACL)}, pages 4487--4497, 2020.

\bibitem{DYY19}
Zihang Dai, Zhilin Yang, Yiming Yang, Jaime~G. Carbonell, Quoc~Viet Le, and
  Ruslan Salakhutdinov.
\newblock Transformer-xl: Attentive language models beyond a fixed-length
  context.
\newblock In {\em Proc. of Conference of the Association for Computational
  Linguistics (ACL)}, pages 2978--2988, 2019.

\bibitem{DCL19}
Jacob Devlin, Ming{-}Wei Chang, Kenton Lee, and Kristina Toutanova.
\newblock {BERT:} pre-training of deep bidirectional transformers for language
  understanding.
\newblock In {\em Proc. Conference of the North American Chapter of the
  Association for Computational Linguistics: Human Language Technologies,
  (NAACL-HLT)}, pages 4171--4186, 2019.

\bibitem{FYZ20}
Jiarui Fang, Yang Yu, Chengduo Zhao, and Jie Zhou.
\newblock Turbotransformers: An efficient {GPU} serving system for transformer
  models.
\newblock {\em CoRR}, abs/2010.05680, 2020.

\bibitem{GHL19}
Linyuan Gong, Di~He, Zhuohan Li, Tao Qin, Liwei Wang, and Tie{-}Yan Liu.
\newblock Efficient training of {BERT} by progressively stacking.
\newblock In {\em Proceedings of International Conference on Machine Learning
  (ICML)}, volume~97, pages 2337--2346, 2019.

\bibitem{GG08}
Kazushige Goto and Robert A. van~de Geijn.
\newblock Anatomy of high-performance matrix multiplication.
\newblock {\em ACM Trans. Math. Softw.}, 34(3), 2008.

\bibitem{HG16}
Dan Hendrycks and Kevin Gimpel.
\newblock {Gaussian Error Linear Units (GELUs)}.
\newblock {\em CoRR}, abs/1606.08415, 2016.

\bibitem{LCG20}
Zhenzhong Lan, Mingda Chen, Sebastian Goodman, Kevin Gimpel, Piyush Sharma, and
  Radu Soricut.
\newblock {ALBERT:} {A} lite {BERT} for self-supervised learning of language
  representations.
\newblock In {\em Proc. of International Conference on Learning
  Representations, (ICLR)}, 2020.

\bibitem{LK20}
Quoc~N. Le and Kip Kaehler.
\newblock {How We Scaled Bert To Serve 1+ Billion Daily Requests on CPUs}.
\newblock
  \url{https://robloxtechblog.com/how-we-scaled-bert-to-serve-1-billion-daily-requests-on-cpus-d99be090db26}.
\newblock Published: 05-27-20, Accessed: 01-06-21.

\bibitem{LOG19}
Yinhan Liu, Myle Ott, Naman Goyal, Jingfei Du, Mandar Joshi, Danqi Chen, Omer
  Levy, Mike Lewis, Luke Zettlemoyer, and Veselin Stoyanov.
\newblock Roberta: {A} robustly optimized {BERT} pretraining approach.
\newblock {\em CoRR}, abs/1907.11692, 2019.

\bibitem{LWY19}
Yizhi Liu, Yao Wang, Ruofei Yu, Mu~Li, Vin Sharma, and Yida Wang.
\newblock Optimizing {CNN} model inference on cpus.
\newblock In {\em Proc. of {USENIX} Annual Technical Conference (ATC)}, pages
  1025--1040, 2019.

\bibitem{Nay19}
Pandu Nayak.
\newblock {Understanding searches better than ever before}.
\newblock
  \url{https://blog.google/products/search/search-language-understanding-bert/}.
\newblock Published: 10-25-19, Accessed: 01-06-21.

\bibitem{NYZ20}
Emma Ning, Nathan Yan, Jeffrey Zhu, and Jason Li.
\newblock Microsoft open sources breakthrough optimizations for transformer
  inference on gpu and cpu.
\newblock
  \url{https://cloudblogs.microsoft.com/opensource/2020/01/21/microsoft-onnx-open-source-optimizations-transformer-inference-gpu-cpu/}.
\newblock Published: 01-20-20, Accessed: 01-06-21.

\bibitem{harvard-nlp}
Harvard NLP.
\newblock {The Annotated Transformer}.
\newblock \url{https://nlp.seas.harvard.edu/2018/04/03/attention.html}.
\newblock Accessed: 01-07-21.

\bibitem{onednn-blas}
oneAPI.
\newblock {BLAS functions}.
\newblock
  \url{https://oneapi-src.github.io/oneDNN/group__dnnl__api__blas.html}.
\newblock Accessed: 01-07-21.

\bibitem{onednn}
oneAPI.
\newblock {oneAPI Deep Neural Network Library (oneDNN)}.
\newblock \url{https://github.com/oneapi-src/oneDNN}.
\newblock Accessed: 01-07-21.

\bibitem{onednn-memory-format}
oneAPI.
\newblock {Understanding Memory Formats}.
\newblock
  \url{https://oneapi-src.github.io/oneDNN/understanding_memory_formats.html}.
\newblock Accessed: 01-07-21.

\bibitem{pytorch}
Pytorch.
\newblock \url{https://github.com/pytorch/pytorch}.
\newblock Accessed: 01-07-21.

\bibitem{pytorch-linear}
Pytorch.
\newblock {Efficient forward pass in nn.Linear}.
\newblock \url{https://github.com/pytorch/pytorch/issues/2159}.
\newblock Accessed: 01-07-21.

\bibitem{pytorch-jit}
Pytorch.
\newblock {Torchscript}.
\newblock \url{https://pytorch.org/docs/stable/jit.html}.
\newblock Accessed: 01-07-21.

\bibitem{pytorch-jit2}
Pytorch.
\newblock {Torchscript for Deployment}.
\newblock
  \url{https://pytorch.org/tutorials/recipes/torchscript_inference.html}.
\newblock Accessed: 01-07-21.

\bibitem{SDC19}
Victor Sanh, Lysandre Debut, Julien Chaumond, and Thomas Wolf.
\newblock Distilbert, a distilled version of {BERT:} smaller, faster, cheaper
  and lighter.
\newblock {\em CoRR}, abs/1910.01108, 2019.

\bibitem{SCG19}
Siqi Sun, Yu~Cheng, Zhe Gan, and Jingjing Liu.
\newblock Patient knowledge distillation for {BERT} model compression.
\newblock In Kentaro Inui, Jing Jiang, Vincent Ng, and Xiaojun Wan, editors,
  {\em Proc. Conference on Empirical Methods in Natural Language Processing},
  pages 4322--4331, 2019.

\bibitem{VSP17}
Ashish Vaswani, Noam Shazeer, Niki Parmar, Jakob Uszkoreit, Llion Jones,
  Aidan~N. Gomez, Lukasz Kaiser, and Illia Polosukhin.
\newblock Attention is all you need.
\newblock In {\em Proc. of Conference on Neural Information Processing Systems
  (NIPS)}, pages 5998--6008, 2017.

\bibitem{WLM20}
Sinong Wang, Belinda~Z. Li, Madian Khabsa, Han Fang, and Hao Ma.
\newblock Linformer: Self-attention with linear complexity.
\newblock {\em CoRR}, abs/2006.04768, 2020.

\bibitem{WWD20}
Wenhui Wang, Furu Wei, Li~Dong, Hangbo Bao, Nan Yang, and Ming Zhou.
\newblock {MiniLM: Deep Self-Attention Distillation for Task-Agnostic
  Compression of Pre-Trained Transformers}.
\newblock In {\em Proc. of Conference on Neural Information Processing Systems
  (NIPS)}, 2020.

\bibitem{WWC20}
Y.~{Wang}, Q.~{Wang}, and X.~{Chu}.
\newblock Energy-efficient inference service of transformer-based deep learning
  models on gpus.
\newblock In {\em IEEE Conferences on Green Computing and Communications
  (GreenCom)}, pages 323--331, 2020.

\bibitem{mcbench}
Yu~Emma Wang.
\newblock {Mille Crepe Bench}: multi-layer performance analysis for deep
  learning frameworks.
\newblock \url{https://github.com/Emma926/mcbench}.
\newblock Accessed: 12-29-20.

\bibitem{WWW19}
Yu~Emma Wang, Carole{-}Jean Wu, Xiaodong Wang, Kim~M. Hazelwood, and David
  Brooks.
\newblock Exploiting parallelism opportunities with deep learning frameworks.
\newblock {\em CoRR}, abs/1908.04705, 2019.

\bibitem{WDS20}
Thomas Wolf, Lysandre Debut, Victor Sanh, Julien Chaumond, Clement Delangue,
  Anthony Moi, Pierric Cistac, Tim Rault, Rémi Louf, Morgan Funtowicz, Joe
  Davison, Sam Shleifer, Patrick von Platen, Clara Ma, Yacine Jernite, Julien
  Plu, Canwen Xu, Teven~Le Scao, Sylvain Gugger, Mariama Drame, Quentin Lhoest,
  and Alexander~M. Rush.
\newblock Transformers: State-of-the-art natural language processing.
\newblock In {\em Proceedings of the 2020 Conference on Empirical Methods in
  Natural Language Processing: System Demonstrations}, pages 38--45, October
  2020.

\bibitem{WBC19}
Carole{-}Jean Wu, David Brooks, Kevin Chen, Douglas Chen, Sy~Choudhury, Marat
  Dukhan, Kim~M. Hazelwood, Eldad Isaac, Yangqing Jia, Bill Jia, Tommer
  Leyvand, Hao Lu, Yang Lu, Lin Qiao, Brandon Reagen, Joe Spisak, Fei Sun,
  Andrew Tulloch, Peter Vajda, Xiaodong Wang, Yanghan Wang, Bram Wasti, Yiming
  Wu, Ran Xian, Sungjoo Yoo, and Peizhao Zhang.
\newblock {Machine Learning at Facebook: Understanding Inference at the Edge}.
\newblock In {\em {IEEE} International Symposium on High Performance Computer
  Architecture (HPCA)}, pages 331--344, 2019.

\bibitem{NYZ19}
Shufan Wu, Tao Lv, Pengxin Yuan, Patric Zhao, Jason Ye, and Haibin Lin.
\newblock {Optimization for BERT Inference Performance on CPU}.
\newblock
  \url{https://medium.com/apache-mxnet/optimization-for-bert-inference-performance-on-cpu-3bb2413d376c}.
\newblock Published: 09-12-19, Accessed: 01-06-21.

\bibitem{XWD20}
Patrick Xia, Shijie Wu, and Benjamin~Van Durme.
\newblock {Which *BERT?} {A} survey organizing contextualized encoders.
\newblock In {\em Proc. of Conference on Empirical Methods in Natural Language
  Processing (EMNLP)}, pages 7516--7533, 2020.

\bibitem{YRH20}
Yang You, Jing Li, Sashank~J. Reddi, Jonathan Hseu, Sanjiv Kumar, Srinadh
  Bhojanapalli, Xiaodan Song, James Demmel, Kurt Keutzer, and Cho{-}Jui Hsieh.
\newblock Large batch optimization for deep learning: Training {BERT} in 76
  minutes.
\newblock In {\em Proc. of International Conference on Learning Representations
  (ICLR)}, 2020.

\bibitem{ZGD20}
Manzil Zaheer, Guru Guruganesh, Kumar~Avinava Dubey, Joshua Ainslie, Chris
  Alberti, Santiago Onta{\~{n}}{\'{o}}n, Philip Pham, Anirudh Ravula, Qifan
  Wang, Li~Yang, and Amr Ahmed.
\newblock Big bird: Transformers for longer sequences.
\newblock In {\em Proc. of Advances in Neural Information Processing Systems
  (NeurIPS)}, 2020.

\bibitem{Zhu19}
Jeffrey Zhu.
\newblock {Bing delivers its largest improvement in search experience using
  Azure GPUs}.
\newblock
  \url{https://azure.microsoft.com/en-us/blog/bing-delivers-its-largest-improvement-in-search-experience-using-azure-gpus/}.
\newblock Published: 11-18-19, Accessed: 01-06-21.

\end{thebibliography}

\end{document}